\documentclass{3dim-article}

\journal{3dim}
\usepackage{url}
\usepackage{hyperref}
\usepackage{cleveref}

\usepackage{algorithm}
\usepackage{algpseudocode}
\usepackage[normalem]{ulem}

\usepackage{lineno}
\usepackage{soul}

\newcounter{tableA}

\articletype{Research Article}

\begin{document}

\title{Event-based Motion-Robust Accurate Shape Estimation for Mixed Reflectance Scenes}

\author{Aniket Dashpute\authormark{1}, Jiazhang Wang\authormark{2,\dag}, James Taylor\authormark{2}, \\ Oliver Cossairt\authormark{3,4}, Ashok Veeraraghavan \authormark{1} and Florian Willomitzer\authormark{2}}

\address{
\authormark{1} Department of Electrical Engineering, Rice University, Houston, TX, 77005\\
\authormark{2} Wyant College of Optical Sciences, University of Arizona, Tuscon, AZ, 85721\\
\authormark{3} Department of Electrical and Computer Engineering, Northwestern University, Evanston, IL, 60208\\
\authormark{4} Department of Computer Science, Northwestern University, Evanston, IL, 60208\\
}
\email{\authormark{\dag}jiazhangwang@arizona.edu } 

\definecolor{myorange}{RGB}{255, 165, 0}
\definecolor{mybrown}{RGB}{155, 78, 20}
\definecolor{mypurple}{RGB}{150, 56, 226}
\definecolor{myblue}{RGB}{10, 50, 230}
\definecolor{myred}{RGB}{160, 10, 10}
\definecolor{mygreen}{RGB}{60, 160, 60}
\definecolor{mydarkgreen}{RGB}{30, 110, 30}
\definecolor{myskyblue}{RGB}{31, 119, 180}
\definecolor{mymagenta}{RGB}{139, 0, 139}
\definecolor{mycyan}{RGB}{0, 100, 100}


\newcommand{\greentick}{{\color{mygreen}{\ding{51}}}}
\newcommand{\redcross}{{\color{red}{\ding{55}}}}

\definecolor{mylightgreen}{RGB}{180, 255, 180}
\definecolor{mylightyellow}{RGB}{250, 250, 150}
\definecolor{mylightred}{RGB}{255, 200, 200}

\newcommand{\ad}[1]{{\color{red}{[Aniket: #1]}}}
\newcommand{\jz}[1]{{\color{orange}{[Jiazhang: #1]}}}
\newcommand{\james}[1]{{\color{myblue}{[James: #1]}}}
\newcommand{\florian}[1]{{\color{mygreen}{\textbf{[Florian: #1]}}}}
\newcommand{\ashok}[1]{{\color{mydarkgreen}{[Ashok: #1]}}}
\newcommand{\ollie}[1]{{\color{mypurple}{[Ollie: #1]}}}
\newcommand{\reb}[1]{{\color{myred}{[Rebuttal: #1]}}}
\newcommand{\delete}[1]{{\color{myred}{[DELETE: #1]}}}

\newcommand{\todo}[1]{{\color{myorange}{[TODO: #1]}}}
\newcommand{\help}[1]{{\color{myred}{[HELP: #1]}}}

\newcommand{\blue}[1]{{\color{myskyblue}{#1}}}
\newcommand{\orange}[1]{{\color{myorange}{#1}}}
\newcommand{\green}[1]{{\color{mygreen}{#1}}}
\newcommand{\red}[1]{{\color{myred}{#1}}}
\newcommand{\purple}[1]{{\color{mypurple}{#1}}}

\newcommand{\beginsupplement}{%
    \setcounter{table}{0}
    \renewcommand{\thetable}{S\arabic{table}}%
    \setcounter{figure}{0}
    \renewcommand{\thefigure}{S\arabic{figure}}%
    \setcounter{section}{0}
    \renewcommand{\thesection}{S\arabic{section}}%
}

\begin{abstract}

Event-based structured light systems have recently been introduced as an exciting alternative to conventional frame-based triangulation systems for the 3D measurements of diffuse surfaces. Important benefits include the fast capture speed and the high dynamic range provided by the event camera - albeit at the cost of lower data quality. 
So far, both low-accuracy event-based and high-accuracy frame-based 3D imaging systems are tailored to a specific surface type, such as diffuse or specular, and can not be used for a broader class of object surfaces (``mixed reflectance scenes"). 
In this work, we present a novel event-based structured light system that enables fast 3D imaging of mixed reflectance scenes with high accuracy.
On the captured events, we use epipolar constraints that
intrinsically enable decomposing the measured reflections into diffuse, two-bounce specular, and other multi-bounce reflections.
The diffuse surfaces in the scene are reconstructed using triangulation.
Then, the reconstructed diffuse scene parts are leveraged as a ``display" to evaluate the specular scene parts via deflectometry.
This novel procedure allows us to use the entire scene as a virtual screen, using only a scanning laser and an event camera.
The resulting system achieves \st{fast and} motion-robust (14Hz) reconstructions of mixed reflectance scenes with $<600 \mu m$ depth error.
Moreover, we introduce an ``ultrafast" capture mode (250Hz) for the 3D measurement of diffuse scenes.\\

\end{abstract}


\section{Introduction}\label{sec:introduction}

\begin{figure}
  \includegraphics[width=\textwidth]{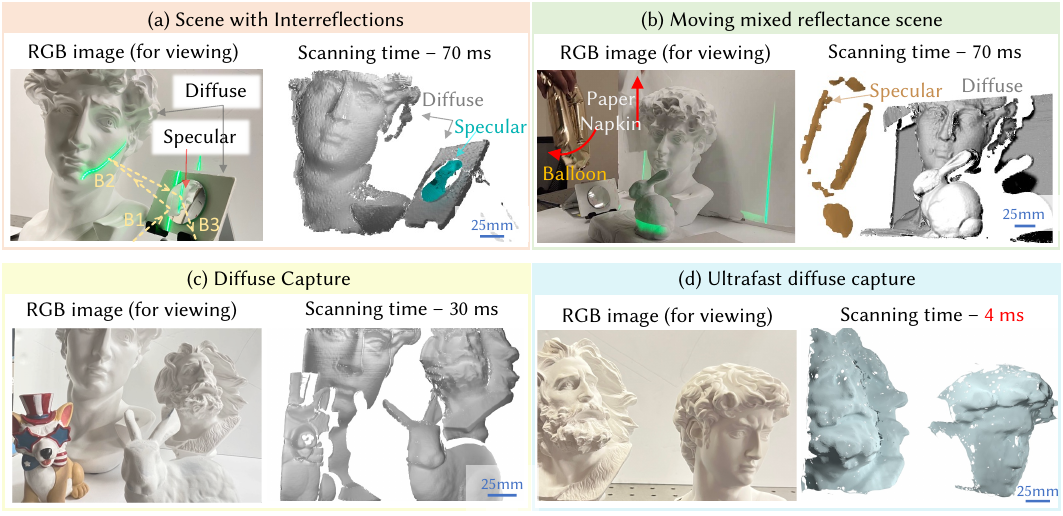}
  \caption{\textbf{We demonstrate a generalized event-based 3D scanning procedure}
  that can simultaneously estimate the shape of diffuse, specular, or partially specular objects with high accuracy (depth error $< 600 \mu m$ 
  Most state-of-the-art 3D scanning methods specialize in estimating the shape of purely diffuse or specular objects. In contrast, our method does not assume any prior knowledge about the reflectance properties or geometry of the scene.
  Event cameras allow fast, high dynamic range captures, resulting in motion-robust measurements of different types of scenes with high accuracy.
  We show high-quality reconstructions of scenes that would be very challenging to reconstruct with conventional methods, such as scenes with interreflections (a) (B1-B3 denote the multiple bounces when light travels from projector to camera), or a mixed reflectance scene (b) in motion (red arrows denote motion: balloon is rotated and napkin is rapidly lifted, see videos \href{https://drive.google.com/drive/folders/1l-5CNM5QqyrP5nZwgZ8e4urpwg04lov8?usp=sharing}{\underline{here}} ).
  We also show measurements of purely diffuse scenes (c), displaying data quality that, to our knowledge, has not been previously achieved with event-based structured light scanners. 
  Moreover, we demonstrate high-quality reconstructions for fast scans with scanning times as low as $4ms$ (d).
  In \Cref{fig:teaser_remaining} we compare our results against state-of-the-art shape estimation techniques and commercial depth cameras like ESL \cite{muglikar2021esl}, Ensemble Codes \cite{gupta2011structured},
  Intel RealSense\cite{intelrealsenseDepthCamera} and Microsoft Kinect\cite{bamji20140}.
  }
  \label{fig:teaser}
\end{figure}

Shape estimation is an important task in computer vision and optical 3D metrology pertinent to multiple fields like robotics, industrial inspection, autonomous navigation, augmented/mixed reality (AR/MR), and even consumer applications like face unlocking on smartphones. Many light transport phenomena occur in a scene with different objects and light sources, the most common being reflection, transmission, emission, and scattering. Depending on the reflectance properties of an object's surface, some of these are more dominant than others. 

Real-world scenes contain a mix of different objects with various surface types, such as highly specular, diffuse, and partially specular (shiny) object surfaces. We call such scenes \textit{mixed reflectance} scenes.
Most state-of-the-art shape estimation methods are tailored toward imaging surfaces of specific reflectance properties like purely diffuse or purely specular. These techniques cannot be used in the wild on mixed reflectance scenes. 

Projecting light (structured or unstructured) onto a mixed reflectance scene can lead to specular reflections, inter-reflections, and subsurface scattering effects. Coding the projected light patterns can help reduce the impact on diffuse object reconstruction \cite{gupta2011structured, o2015homogeneous, gupta2015phasor, nayar2012diffuse}. We can also use the geometry of the scene (epipolar constraints) to separate out the direct, diffuse components from indirect illumination \cite{o20143d, criminisi2005extracting, yang2023sepi}.
Sparse-projection techniques (like point or line projection) are effective in applying these constraints as they limit the inter-reflections and the resulting signal ambiguities in the scene.
For frame-based camera systems, this comes with the drawback of longer capture time, leading to motion artifacts.
We define systems in which the reconstructed shapes or depths do not suffer from motion artifacts as \textit{motion-robust}.
Most 3D imaging systems that combine sparse-projection techniques with frame-based cameras are not motion robust, due to the extended time it takes to capture a dense scene representation via sparse projection.

Event cameras \cite{gallego2020event, lichtsteiner2008128, brandli2014240, posch2010qvga} are differential cameras that can mitigate these drawbacks as they have extremely fast read-out for sparse changes in the scene. Event-based triangulation has been demonstrated for reconstructing \textit{diffuse scenes} \cite{matsuda2015mc3d, muglikar2021esl, wang2020joint, huang2021high} in an acquisition time as low as $16ms$. 
In this work, we tackle the challenging problem of \textit{capturing mixed reflectance scenes}, i.e., scenes that simultaneously contain object surfaces with diffuse, specular, and "shiny" reflectance properties.
We use event-based scanning system, separate the diffuse and specular components using epipolar constraints, and reconstruct the scene in a two-step post-capture process. We first evaluate the diffuse surfaces via triangulation. We then use these diffuse surfaces as a secondary illuminating source (a virtual screen) and observe the reflection of this virtual screen over all the specular surfaces in the scene. This allows the evaluation of all specular surfaces via deflectometry, a standard optical-metrology technique for ultra-high-precision reconstruction of lenses and technical parts \cite{knauer2004phase, huang2018review, faber2012deflectometry}.

Our novel approach combines the benefits of high-accuracy triangulation and deflectometry approaches, has high speed (14Hz up to 250Hz), and high pixel count (1 megapixel), which enables the motion-robust measurement of mixed reflectance scenes.
To the best of our knowledge, a robust solution to this problem with high accuracy has not been demonstrated so far.
Our work presents the first step towards a novel concept that could enable a new wave of 3D sensors.
Imagine the measurement of real-life scenes like the interior of a car or a living room. These scenes have a mix of specular objects like mirrors and shiny displays and diffuse objects like walls, cloth, and furniture.
A 3D sensor that allows for the shape estimation of such scenes could potentially enable novel high-accuracy imaging applications in VR/MR, enabling users to precisely grasp small objects like a pen or button, or assist small robots or drones to navigate through the most crowded and complicated environments.
It can also help in medical applications like enabling surgical machines to make accurate cuts even on shiny and moving objects such as an open heart. Such scene-independent, high-speed, and precise 3D sensing has a broader impact, and stakeholders from different fields can profit from this.

The contributions of our work include:
\begin{itemize}
    \item \textbf{3D Imaging solution that can accommodate different surface reflectance properties:} A high-resolution 3D imaging system that can reconstruct the shape of diffuse, specular, and partially specular surfaces simultaneously.
    \item \textbf{Motion and ambient light robust} system, imaging at 14Hz for mixed reflectance scenes and up to 250Hz for diffuse scenes,
    being robust to any changes in ambient lighting.
    \item \textbf{Diffuse surfaces as virtual screens:} For the first time, we show that diffuse surfaces in a scene can be repurposed as deflectometry screens to reconstruct specular shapes.
    \item \textbf{Event Deflectometry:} For the first time, we demonstrate deflectometry measurements using event cameras.
    \item \textbf{Novel dual-scanning} laser projector for fast sparse scanning to enable direct-indirect illumination separation and specular shape estimation via deflectometry.
    \item \textbf{High accuracy:} Quantitative analysis shows that our method performs with a depth error $< 600 \mu m$ much lower than comparable state-of-the-art sensors.
    \item \textbf{Optimization-based deflectometry reconstruction:} We also introduce a novel algorithm for deflectometry shape reconstruction for a single camera-projector system.
\end{itemize}

\begin{table}
    \refstepcounter{tableA}
    \centering
    \hspace{-10pt}
    \begin{tabular}{|c|c|c|c|} \hline 
         \textbf{Method}&  \textbf{Exposure $\tau$} &  \textbf{Depth Error}&  \textbf{Scenes}\\ \hline 
         Ensemble Codes\textsuperscript{\cite{gupta2011structured}} &  \cellcolor{mylightred}50 frames&  \cellcolor{mylightgreen}  $\sim$1.4 mm  $^\dagger$&  \cellcolor{mylightred}D\\ \hline 
         Kinect v2 ToF \textsuperscript{\cite{bamji20140}}&  \cellcolor{mylightgreen}33 ms &  \cellcolor{mylightred}$\sim$2 mm $^\dagger$&  \cellcolor{mylightred}D\\ \hline 
         Intel Realsense \textsuperscript{\cite{intelrealsenseDepthCamera}}&  \cellcolor{mylightgreen}11-33 ms &  \cellcolor{mylightred}$\sim$1 cm $^\dagger$& \cellcolor{mylightyellow}D+PS\\ \hline 
         MC3D\textsuperscript{\cite{matsuda2015mc3d}} &  \cellcolor{mylightgreen}16 ms &  \cellcolor{mylightred}$\sim$8 mm $^\dagger$&  \cellcolor{mylightred}D\\ \hline 
         ESL\textsuperscript{\cite{muglikar2021esl}} &  \cellcolor{mylightgreen}16 ms & \cellcolor{mylightred} {$\lesssim$ 6 mm *}&
         \cellcolor{mylightred}D \\ \hline 
         \textbf{Ours} &  \cellcolor{mylightgreen}4-70 ms & \cellcolor{mylightgreen} \textless {0.6 mm $@ \tau = 16 ms$} & \cellcolor{mylightgreen}D+S+PS \\ \hline
    \end{tabular}
    \caption{\textbf{Comparison of our method against state-of-the-art and commercial sensors.} Our method performs better in terms of motion robustness (fast capture), low depth error, and types of scenes that can be imaged. (\textit{Notations used}: \textbf{D}: Diffuse surfaces, \textbf{S}: Specular surfaces, \textbf{PS}: partially specular surfaces).
    $^\dagger$ Values for comparing methods are taken from respective data sheets or publications.
    * Implemented by us with the same event camera, baseline, and triangulation angle as ours.
     A comprehensive analytical comparison would require considering additional system parameters (such as measurement volume, stereo baseline, or ability to measure color). Additional details for hardware comparisons are provided in Supplementary Table S1.}
    \label{tab:Comparison_\thetableA}
\end{table}
\section{Background and Related Work} \label{sec:background}

\begin{figure*}
    \centering
    \includegraphics[width=\linewidth]{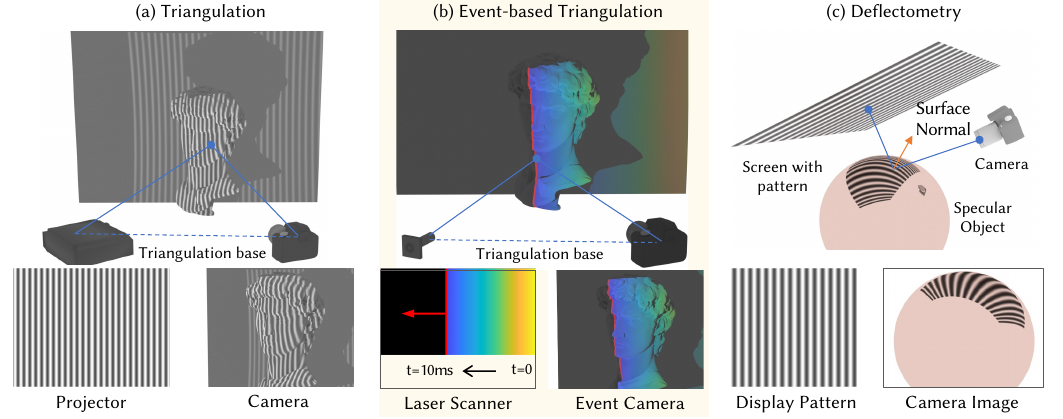}
    \caption{\textbf{3D Imaging Principles.} A camera and a projector form an active triangulation system (a). The projector can be treated as an inverse camera and the two together form a stereo vision system. By replacing the projector with a laser scanner and the camera with an event sensor, we get an event-based triangulation system (b). Triangulation systems work for the measurement of diffuse surfaces but fail for specular surfaces. Deflectometry is one of the special techniques applicable to the shape estimation of specular surfaces (c). In conventional deflectometry setups, an extended screen whose position is known with respect to the camera is used. The screen displays known patterns and the camera observes the reflection of these patterns over the specular object surface. From the deformation of the patterns in the camera image, the surface normals (and later the shape via optimization) are calculated.}
    \label{fig:Background}
\end{figure*}

Most state-of-the-art shape estimation techniques can be broadly classified into three categories \cite{hausler2022reflections}:
(a)~triangulation-based principles (including passive and active stereo, light field imaging, depth from focus, or ``structured light''),  (b) reflectance-based principles that measure the surface gradient (such as photometric stereo, deflectometry, or shape from polarization), and (c)~principles that measure the ``time of flight'' (ToF) or travel distance of light (including ToF cameras, LiDAR, OCT, or interferometry). We discuss different measurement principles from these categories with respect to their ability to measure diffuse and /or specular object surfaces. Special attention is given to principles that can measure macroscopic objects with high accuracy.

\subsection{Shape measurement of diffuse surfaces} \label{sec:DiffuseShape}

Diffuse object surfaces scatter an incident light ray in multiple directions, meaning that a light signal reaching the surface can be ``seen'' and measured from multiple angles outside the angle of direct reflection. This is beneficial for 3D imaging, as surfaces with a wide variety of orientations can be measured with point light sources or projectors, i.e., sources where the light only comes from one single point. A special case of diffuse scattering is Lambertian scattering, where the surface appears equally bright from all directions and the brightness only depends on the angle of incidence of the light ray. This assumption forms the basis for \textit{Photometric Stereo} \cite{woodham1979photometric} or \textit{Shape from Shading} \cite{zhang1999shape}.

Similar brightness from multiple viewpoints is also useful for active or passive \textit{triangulation} principles:
Two cameras whose positions relative to each other are known, observe a point on the diffuse object surface. This forms a triangle with known angles and baseline, and the depth can be calculated by simple ray intersection. In active triangulation, patterns are projected onto the (possibly texture-less) object surface to obtain better correspondence between the cameras. The projector can also be calibrated as an inverse camera and can replace the second camera. \Cref{fig:Background}(a) shows an active triangulation system that uses a calibrated projector with one camera. The pattern projected onto the object's surface looks deformed in the camera picture. The 3D shape is calculated from this deformation. The simplest form of active triangulation is (laser) point raster scanning: A single point is scanned over the object surface. For each time instance (e.g., camera image), the camera only observes one bright point, which directly delivers projector-camera correspondence free from any ambiguity.
Over the years, different forms of active triangulation have been invented and developed. Many of them differ in the kind of patterns that are projected, which leads to benefits and drawbacks with respect to measurement speed, surface feature resolution, and size of the measurement volume \cite{geng2011structured}. Examples include phase-measuring triangulation \cite{zuo2018phase,srinivasan1984automated}, Fourier transform profilometry \cite{takeda1983fourier}, dense single-shot line triangulation \cite{willomitzer2017single, willomitzer2013flying}, single-photon structured light \cite{sundar2022single}, fast triangulation using speckle patterns\cite{Schaffer:10} or hybrid methods \cite{mirdehghan2018optimal}.

In general, it can be said that denser patterns deliver higher point cloud densities, but are in turn more susceptible to correspondence ambiguities.
Such ambiguities are commonly resolved by exploiting a temporal sequence of exposures (potentially not motion-robust), by using sophisticated spatial pattern codifications (possibly losing some high-frequency details of the surface), or by exploiting prior knowledge about the scene, which includes recent learning-based methods \cite{zuo2022deep}.
Carefully implemented triangulation systems can measure diffuse surfaces of macroscopic objects with an impressive precision of
hundreds or even tens of $\mu m$ \cite{cyberopitcsDSSeries}.
However, triangulation principles also have a weak spot: they fail to measure specular surfaces.

\subsection{Shape measurement of specular surfaces} \label{sec:SpecularShape}

Specular surfaces like mirrors reflect an incident light ray only in one direction. This direction depends on the angle of incidence and the surface normal. For measurement principles like triangulation, ToF, or photometric stereo, the light only comes from one single point (such as the nodal point of a projector or a point light source).
This means that a significant portion of reflected light rays does not find its way back into the camera, which makes specular objects quasi ``invisible'' for these principles. It should be noted, however,  that photometric stereo-based setups for the measurement of specular surfaces exist \cite{ikeuchi1981determining, solomon1996extracting}. Nevertheless, these setups typically would require a large number of point light sources to be able to measure a high number of surface points, i.e., to achieve a good \textit{coverage} of the surface.
Another technique known as shape-from-polarization enables the measurement of specular surfaces \cite{rahmann2001reconstruction}. It has been recently augmented with event sensors to enhance its operational speed \cite{muglikar2023event}. Nevertheless, the reliance on orthographic projection and unpolarized light limits its ability to achieve high levels of accuracy and practical usage.

An easy and intuitive way to increase the coverage for the active 3D measurement of specular surfaces is to increase the \textit{spatial (and angular) extent of the light source}. This is the basic idea behind \textit{deflectometry}, which is a well-known measurement principle in optical metrology \cite{knauer2004phase, huang2018review, faber2012deflectometry,burke2023deflectometry}. 
Deflectometry setups commonly use a display (e.g., LCD screen) as a light source. The reflection of this display at the specular surface is observed with a camera (\Cref{fig:Background}(c)). From the deformation of the (known) display pattern in the camera image, the normal map of the specular object can be calculated. Eventually, the object shape can be retrieved by gradient integration methods like the Frankot-Chellappa algorithm \cite{frankot1988method}, or iterative surface integration methods \cite{huang2012improvement, huang2015comparison}.
Well-calibrated deflectometry setups can reach sub-micron depth resolution, which is the reason why deflectometry is widely used for industrial surface inspection of optical components \cite{shimizu2021insight, knauer2004phase}, to measure car bodies \cite{hofer2016infrared},
for cultural heritage \cite{willomitzer2020hand}
or for eye surface measurement and tracking \cite{liang2016single, wang2023accurate, wang2023optimization}.
Deflectometry is classified as an active imaging method because it uses an external display to project patterns on specular surfaces.
Passive specular reconstruction using environment maps has been shown before by
Oren and Nayar \cite{oren1997theory}, Nishino and Nayar \cite{nishino2004eyes},
and more recently by Tiwary et al. \cite{glossyobjects2022} using neural rendering techniques.
These passive methods commonly require the scene to be sufficiently ``structured", leading to low-quality results in dark or homogeneous-looking environments.

In deflectometry, surface coverage for the measurement of specular objects is dependent on the object shape (e.g., convex, concave), as well as the size and standoff distance of the display. Increasing the coverage for complicated specular objects without using an elaborate dome arrangement is still a big problem for current state-of-the-art methods. A solution to this problem is one of the contributions of our work.

\subsection{Shape measurement of mixed reflectance scenes}

All methods discussed above are tailored to the reflectance type of the target objects, i.e., they work best for purely diffuse or purely specular surfaces. General real-world scenes, however, are composed of a mix of diffuse and specular object surfaces.
Many real-world object surfaces don't have purely specular or diffuse BRDFs (Bidirectional Reflectance Distribution Functions) but exhibit partially specular reflectance behavior.
Those objects are very hard to measure with conventional methods, and the results commonly leave much room for improvement. One reason is the vastly different signal return for the partially specular surfaces (very bright around direct reflection direction, very dim outside), which cannot be properly captured by the dynamic range of conventional cameras and leads to very different signal-to-noise ratios (SNRs). For the remainder of this paper, we refer to scenes composed of diffuse, specular, and partially specular (shiny) objects as \textit{mixed reflectance scenes}.
Previous work has attempted to estimate the shape of mixed reflectance scenes, e.g., by further extending diffuse triangulation-based methods.
Projecting structured light onto such a scene leads to unwanted inter-reflections due to specularities and scattering effects. One way to improve the measurement outcome is to filter out these inter-reflections (the \textit{indirect part}) from the diffuse components (the \textit{direct part}\footnote{which also includes the special case of direct specular component (\Cref{fig:AlgorithmExplanation} Case 3)} ). Gupta et al. \cite{gupta2011structured} design structured light patterns with high spatial frequencies to eliminate the indirect illumination effects in triangulation measurement.

This can also be done by exploiting epipolar constraints. The epipolar geometry \cite{szeliski2022computer} restricts the direct component reflection to a specific region of pixels on the image plane, and anything outside this region can be classified as indirect illumination.
O'Toole et al. \cite{o20143d} physically separate the components using a rolling shutter camera with a rolling aperture projector. Extracting the direct components allows the method to get the shape of the diffuse scene components (including the special case of specular direct (\Cref{fig:AlgorithmExplanation} Case 3)). However, indirect reflections are rejected, and the shape of specular objects generally cannot be recovered.
Tsai et al. \cite{tsai2016shape} provide a theoretical foundation for recovering shape and reflectance from two-bounce light transients in systems where the path lengths and light transport can be measured (such as a femto-second laser/streak camera system \cite{velten2012recovering} or laser diode/ToF camera system \cite{heide2014diffuse}).

Nayar et al. \cite{nayar2012diffuse} use diffused structured light to recover shape of mixed reflectance scenes - particularly diffuse and challenging partially specular (shiny) surfaces. However, the method still fails
to measure very shiny and purely specular surfaces with high accuracy.

Another class of techniques combines deflectometry for specular shape measurement and structured light triangulation for diffuse shape measurement \cite{huang2011study, liu20203d}. Their setup consists of a projector-camera system for triangulation and an LCD screen-camera system for deflectometry. Due to the combination of two principles, the method can measure diffuse and specular objects at the same time, but some fundamental problems discussed above remain: The coverage of specular objects is still dependent on the screen size, and the use of large screens becomes less practical with increasing scene size.
Karami et al. \cite{karami2022combining} have similarly combined photogrammetry and photometric stereo to capture specular and diffuse parts but suffer from inherent bulky setup limitations imposed by photometric stereo.

\section{Methods} \label{sec:Methods}

\begin{figure*}
    \centering
    \includegraphics[width=\linewidth]{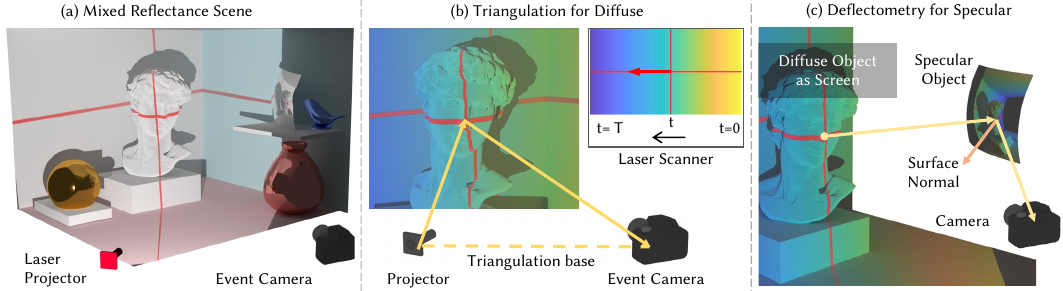}
    \caption{\textbf{Shape estimation of a mixed reflectance scene.} We image a mixed reflectance scene (a), consisting of a variety of diffuse, partially specular (shiny), and specular object surfaces. We use a laser scanner to scan the scene. We then separate the diffuse and specular components from the scan. For the diffuse components of the scene, we estimate shape using event-based triangulation (b). These diffuse components now act as a screen for the deflectometry setup and we can accordingly evaluate the shape of the specular objects (c).}
    \label{fig:OurApproach}
\end{figure*}

In our proposed approach, we combine triangulation and deflectometry to \textit{simultaneously scan all surface types} in a mixed reflectance scene.
Our setup consists only of an event camera and a scanning laser (laser diode + galvo). The basic pipeline is shown in \Cref{fig:OurApproach}.
We scan the mixed reflectance scene with the laser and separate the diffuse and specular scene contents using epipolar geometry. Eventually, we evaluate the 3D shape of the diffuse scene contents via triangulation. The specular scene contents are subsequently evaluated via deflectometry. However, instead of using an external screen for the deflectometry evaluation, we repurpose the just-evaluated 3D coordinates of the diffuse scene contents as a screen. One obvious benefit of this procedure is the low hardware and calibration effort: Only one camera and one scanning laser are needed, and no external screen. However, a more important benefit concerns the coverage of the specular objects in the scene: since \textit{everything around is a screen}, the coverage restriction depends only on the scanned scene contents.
All diffuse objects in a room can be used as a screen!

\subsection{Event-based 3D Scanning} \label{sec:EventTriangulation}

Event cameras are biologically inspired vision sensors that output a stream of events instead of complete image frames. A pixel is triggered and an \textit{event} is read out only if the log of its intensity value $I$ changes by more than a set threshold $\epsilon$ (controlled by the bias parameters of the camera) in an interval $\Delta t$ (dictated by a number of factors including bias parameters and rate of intensity change at that pixel).
A collection of these events forms the event stream.
Each event $e_{ijt}$ in the stream is asynchronously transmitted as a tuple of pixel coordinate $(i, j)$ and the associated timestamp $t$, along with the polarity $p$ (positive or negative) of the intensity change,

\begin{equation}
    e_{ijt} = \left(i, j, t, p\right),\text{ }s.t. \text{ } \mid \log(I(i,j,t)) - \log(I(i,j,t-\Delta t)) \mid > \epsilon.
    \label{eq:EventDef}
\end{equation}

The sparse read-out scheme of the event cameras allows imaging fast-changing scenes with latencies down to the order of 10 $\mu s$, depending on the sparsity of the events \cite{gallego2020event}.
The differential nature of event cameras also drastically increases their dynamic range and robustness to ambient light with respect to frame-based cameras. In contrast, frame-based cameras require adjustment of exposure times for different lighting conditions and surface types.

So far, 3D imaging with event cameras has been mainly limited to \textit{event-based triangulation} for diffuse surfaces. The basic idea is that \textit{multi-shot} triangulation principles with \textit{sparse} patterns (such as point raster scanning or line sweeping using lasers) can be significantly sped up by using a low latency event camera \cite{matsuda2015mc3d, muglikar2021esl, wang2020joint, huang2021high, morgenstern2023x}.  \Cref{fig:Background}(b) depicts the basic principle. A correspondence between the projector and the camera pixels is obtained through the detected timestamps. This correspondence is then used to obtain a depth value for each camera pixel via triangulation.

Our introduced method uses a combination of two perpendicular line lasers to scan the scene by alternating between horizontal and vertical sweeps.
We use these measurements to \textit{computationally synthesize an equivalent point scanning system},
where each point position is obtained from the intersection of the scanned vertical and horizontal lines (see \Cref{fig:DualScanProjector}).
This procedure provides a significant speed advantage over single-point raster scanning. Assuming a square (two-dimensional) scan area whose side‐length sweep takes time $T$, raster scanning requires $T^2$ total time, whereas dual orthogonal sweeps take only $2T$ (see Supplementary Section S1.1 for more details).
We note that a sweep in two directions is not necessary for triangulation on purely diffuse scenes, as a single line sweep in conjunction with epipolar geometry can be used for shape estimation. However, there is no epipolar constraint in deflectometry, meaning that a point scanning equivalent (sweep in two directions) is essential for epipolar separation (\Cref{sec:DiffSpecSep}) and its subsequent deflectometry evaluation (\Cref{sec:EverythingIsScreen}).

The high dynamic range of the event camera also improves the measurement of partially specular surfaces. For example, a low-intensity diffuse reflection from the surface is still sufficient to evaluate this surface in ``triangulation mode''. In the following subsections, we give a detailed explanation of the steps involved in our approach.

\subsection{Diffuse-Specular Separation} \label{sec:DiffSpecSep}

\begin{figure}
    \centering
    \includegraphics[width=\linewidth]{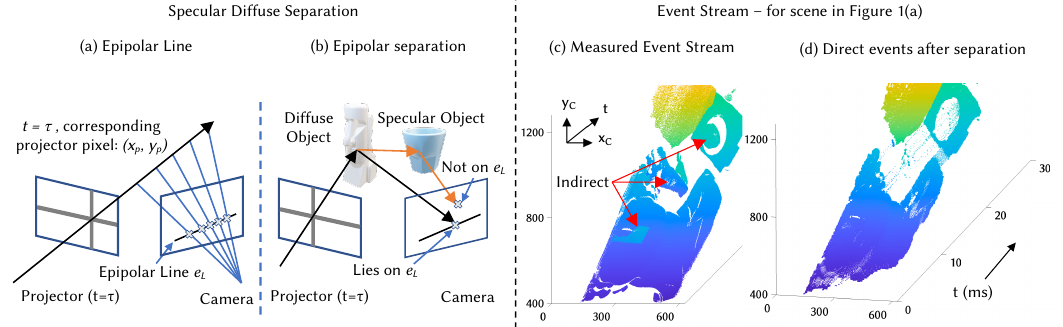}
    \caption{\textbf{Separation of Direct and Indirect components using Epipolar constraints.} When we image a light ray being projected from a projector pixel $(x_P, y_P)$ at different depths in object space, all respective point images will lie on a single line on the camera image plane, This line is the epipolar line $e_L$ as shown in (a). This means that single-bounce reflections from this light ray always land on $e_L$. Light rays that undergo multi-bounce reflections do not typically land on $e_L$, as seen in (b). This information is used in our work to classify direct reflections, which lie on $e_L$, and indirect reflections that don't (\Cref{fig:AlgorithmExplanation} shows all the such cases of reflections and how they are handled in our method).
    In (c), we show the measured \textit{event stream} for the measurement from \Cref{fig:teaser}(a) as a $x, y, t$ point cloud (c). This has direct reflection components (only diffuse direct in this case) as well as indirect components (two-bounce and multi-bounce reflections). The direct component after epipolar separation is shown in (d). Notations used: \textbf{(x$_C$, y$_C$):} event camera pixel, \textbf{t:} timestamp in \textit{ms}.
    }
    \label{fig:SpatioTemporalEpipolar}
\end{figure}

\begin{figure}
    \input{figures_tex/Fig_algorithm_explanation}
    \label{fig:AlgorithmExplanation}
\end{figure}

\begin{algorithm}
    \caption{Correspondence from dual-scanning and classification of reflections}
\begin{algorithmic}

\For {$x_P =1,2,\ldots,w_P$}
    \For {$y_P =1,2,\ldots,h_P$}
        \State $t_x = ({x_P}/{w_P}) \times T_{scan}$
        \State $t_y = ({y_P}/{h_P}) \times T_{scan}$
        \\
        \State \textbf{I. Get events corresponding to $(x_P, y_P)$:}
        
        \State $E_x = E_{x-events}(t_x - \Delta t : t_x + \Delta t)$ \Comment{events $(x_C, y_C, t)$ corresponding to the vertical scan line of $x_P$}
        \State $E_y = E_{y-events}(t_y - \Delta t : t_y + \Delta t)$ \Comment{for horizontal line scan associated with $y_P$}

        \State \(X^x_{c} \;=\; \{\,(x_c, y_c)\mid (x_c, y_c)\,\in\,E_x\}\)
        \State \(X^y_{c} \;=\; \{\,(x_c, y_c)\mid (x_c, y_c)\,\in\,E_y\}\)
        \Comment{each set collects all \((x_c, y_c)\) pairs from \(E_x\) and \(E_y\), respectively.}
        \\
        \State \textbf{II. Intersection of the two scan lines gives the correspondence:}
        \State \( X^{xy}_{c} \;=\; \{\,(x_c, y_c)\;\mid\;(x_c, y_c)\in E_x \;\cap\; E_y\} \)
        \State $(X^1_c, X^2_c, \ldots, X^k_c) = \textbf{cluster}(X^{xy}_c, k)$
        \Comment{Make k clusters to get mean position of the scan line intersections}

        \\
        \State \textbf{III. Classification based on epipolar constraints:}
        \State \textbf{Let} \(X^{\mathrm{epi}}_c\) \textbf{be the unique point} \(X^i_c\) \textbf{such that} \(X^i_c \in e_L.\)
        \Comment{$e_L$ is the epipolar line for $(x_P, y_P)$}

        \If{$X^{\mathrm{epi}}_c$ \textbf{exists}} 
        \Comment{True if a unique epipolar intersection was found}
            \State $X_\mathrm{direct}(x_P, y_P) \gets X^{\mathrm{epi}}_c$
            \State \text{direct\_mask}$(x_P, y_P)$ = 1

            \If{$\textbf{more intersection(s) } X^{\mathrm{i}}_c \neq X^{\mathrm{epi}}_c$  \textbf{ exist}} 
            \State $X_\mathrm{indirect}(x_P, y_P) \gets X^{\mathrm{i}}_c$
            \Comment{We restrict this to only one more intersection, but more could be included}
            \EndIf
        \EndIf

    \EndFor
\EndFor
\Comment{We now have all the direct reflections in  $X_\mathrm{direct}$, and all the \textit{corresponding} indirect ones in $X_\mathrm{indirect}$}
\\
\State \textbf{IV. Keep only deflectometry (Case II) indirect reflections:}
\State $X_\mathrm{specular} = (1 - \mathrm{direct\_mask}) \cdot X_\mathrm{indirect}$

\Comment{removes any scattering noise, keeps specular bounces for deflectometry}
\end{algorithmic}
    \label{alg:intersection_classification}
\end{algorithm}

\begin{algorithm}
    \caption{Iterative shape optimization for deflectometry}
\begin{algorithmic}

\State \textbf{Initiate a depth estimate for specular shape}
\State $\mathrm{initialize\ depth\_image}$
\State $\mathrm{corresponding\ screen\_points\ obtained\ from\ Algorithm\ 1 }$
\\
\For {\textbf{iter = 1, 2, $\cdots$, N }}
    \\
    \State \textbf{I. Get 3D surface from depth and camera params K}
    \State $\mathrm{surface\_points} = \mathrm{get\_xyz\_from\_z(depth\_image, K)}$ \Comment{K: camera intrinsics}
    \\
    \State \textbf{II. Get normals from surface: Get two lines (tangents) along the surface and get cross-product}
    \State $\mathrm{obtain}$  \textbf{t$_W$, t$_H$} \Comment{central differences along width and height}
    \State $\mathrm{normals\_surface = \textbf{t$_W$ $\times$ t$_H$}}$

    \\
    \State \textbf{III. Get normals from screen points}
    \State \textbf{rays\_1} $\mathrm{ = ray(surface\_points \rightarrow camera\_node})$
    \Comment{Ray 1 goes from surface point to camera node}
    \State \textbf{rays\_2} $\mathrm{ = ray(surface\_points \rightarrow screen\_points})$
    \Comment{Ray 2 goes from surface point to screen point}
    \State $\mathrm{normals\_correspondence}$ = \textbf{rays\_1} + \textbf{rays\_2}

    \\
    \State \textbf{IV. Loss function: Surface normals should be close to normals from correspondence}
    \State $\mathrm{loss\ Q = cosine\_similarity(normals\_surface, normals\_correspondence) + \lambda\ smoothness(depth\_image)}$

    \\
    \State \textbf{V. Refine depth estimate}
    \State $\mathrm{depth\_image} \leftarrow \mathrm{depth\_image} - \eta \nabla_{\mathrm{depth\_image}} Q$
    \\
\EndFor

\end{algorithmic}

    \label{alg:iterative_shape}
\end{algorithm}

If the measured scene is completely diffuse, every generated event can be used for triangulation. In mixed reflectance scenes, however, we also encounter specular objects for which triangulation cannot be applied. When estimating the shape of those scenes, we use triangulation for the diffuse reflections (\Cref{fig:OurApproach}(b)) and deflectometry for the specular counterpart (\Cref{fig:OurApproach}(c), \Cref{sec:EverythingIsScreen}). This requires us to separate the single-bounce (direct) reflections from multi-bounce (indirect) reflections.

We exploit epipolar constraints for the separation. Epipolar geometry describes the relation between two camera images in a triangulation (stereo vision) system. The relation also applies when one of the cameras is replaced with a projector, which is then treated as an inverse camera. \Cref{fig:SpatioTemporalEpipolar}(a) demonstrates the concept: for a particular projector pixel, the stereo geometry constrains the corresponding point on the camera plane to lie on a specific line - the epipolar line.
These epipolar constraints can only be applied to the direct components of the scene, that is when the light from the projector reaches the camera through a single bounce. This happens for single-bounce diffuse reflections and single-bounce specular reflections that are directly reflected back to the camera center \footnote{This is a special case for specular reflection that can also be evaluated with triangulation.}. For multiple bounces (inter-reflections), this geometric constraint is not valid, as seen in \Cref{fig:SpatioTemporalEpipolar}(b).
In turn, this means that any point in the camera image
that does not lie on its epipolar line must come from a multi-bounce reflection.
We use a threshold of $\pm 2$ pixels for epipolar line classification.

To facilitate fast, efficient, and
accurate scanning, we develop a novel dual-scan laser projector that can be
algorithmically converted to a point raster scanner (\Cref{fig:DualScanProjector}, implementation details in Supplementary Section 1.1.)
This approach is equivalent to lighting up one projector pixel at a time (albeit significantly faster).
For each such projector pixel, we use its epipolar line to separate out the directly and indirectly illuminated pixels.
This idea was used by O'Toole et al. \cite{o20143d} to separate direct/indirect components for triangulation-based reconstruction.
Yang et al. \cite{yang2023sepi}
demonstrated this idea with the use of an event camera but use gray code projection patterns, making their method slow, prohibiting the capture of non-stationary scenes.
O'Toole et al. \cite{o20143d} demonstrated their approach for direct/indirect separation, classifying all specular, sub-surface scattering, and inter-reflection components as ``indirect''.
For our method (see \Cref{fig:AlgorithmExplanation} for various reflection possibilities and \Cref{alg:intersection_classification} their treatment), we additionally need to separate the indirect specular components (to be evaluated with deflectometry) from the other indirect components.
These are a specific set of two-bounce reflections that we need to keep. Specifically, we ensure that the first bounce for these reflections comes from a diffuse surface, and the second one from the specular surface, exactly as seen in \Cref{fig:AlgorithmExplanation}, Case II. This can be done by recovering the first bounce direct component using epipolar separation. The other signal is then its corresponding reflection from the specular surface. This signal is eventually evaluated with deflectometry, where the direct first-bounce component forms the screen for the second-bounce specular surface reflection (see \Cref{alg:intersection_classification} for details) .

\subsubsection*{Measurement of diffuse and partially specular regions. }
After separating the reflection components, we use event-based triangulation on the diffuse parts, as described in \Cref{sec:EventTriangulation}.
Some partially specular objects like a plastic cup show both diffuse and specular reflections. For a particular pixel covering this object, it might get a diffuse reflection for some timestamp and specular for another. In such cases, we classify that pixel as diffuse and reject its specular component. We then use triangulation for its shape estimation as we do for all the diffuse components. We can do this because of the high dynamic range of the event camera, as we still get a good signal if we only evaluate the (sometimes very dim) diffuse component. This is another benefit of using event cameras.

\subsection{Everything Around is a Screen} \label{sec:EverythingIsScreen}

\subsubsection*{Deflectometry for specular scene parts.}
A standard deflectometry system consists of a screen, the specular object to be measured, and the camera. The screen and the camera position are calibrated with respect to each other before the measurement. When measuring the normals (and later the shape) of a specular surface, first the correspondence between screen coordinates and camera pixels is determined. In other words, for each camera pixel that observes the specular object surface, we need to know the 3D coordinate of the corresponding point on the screen that illuminates the specular object. In our proposed approach, we already have those corresponding ``3D screen coordinates'' as they are nothing but the 3D coordinates of the diffuse scene parts and have been calculated in the previous triangulation step (see \Cref{fig:OurApproach}).
Thus \textit{any diffuse surfaces in the scene can be used as a screen for the shape estimation of specular surfaces}. As this removes the necessity of a separate screen for deflectometry, everything around can be treated as a screen!

This approach has several advantages over conventional deflectometry systems. First, no additional ``screen calibration'' is necessary, as the \textit{screen} coordinates are automatically obtained from the triangulation evaluation step. A second advantage is that we can theoretically create arbitrary large screens (consisting potentially of entire rooms) to significantly improve the coverage of specular surfaces. The idea of using everything around as a screen was explored before in the context of display systems, where arbitrarily shaped objects could be measured and used as displays from given user perspectives \cite{raskar1998office, raskar2002projector}. 
Parallels to non-line-of-sight problems could be drawn as well, as respective sensor concepts exploit relay walls or even different objects in the scene as secondary sources of reflection
\cite{xin2019theory, faccio2020non, willomitzer2021fast, gu2023fast}.

Getting the exact correspondence between the camera and screen coordinates is necessary for deflectometry. This would not be possible if we just used a line-scanning projector, where a specular surface pixel would ambiguously map to several diffuse screen coordinates. As discussed before, point scanning is especially helpful in this case as the number of signal ambiguities is significantly decreased, and specular surfaces can be mapped to their corresponding diffuse screen points.
Getting this accurate correspondence gives us an estimate of the surface normals (\Cref{fig:OurApproach}(c)).
To calculate the shape of the specular shapes, we set up a joint optimization problem that infers specular shape (depth) and normals from the measured correspondence map. In this iterative process, the algorithm reduces the cosine similarity loss between the normals obtained from the correspondence map (formed by surface $\rightarrow$ screen ray and surface $\rightarrow$ camera ray) and the normals obtained from the calculated surface (calculated via the gradient of the obtained shape), until it converges. The optimized depth from this procedure is the resultant specular shape. This process is described in \Cref{alg:iterative_shape}.

We combine the diffuse and specular surface reconstructions to get the shape estimate of a mixed reflectance scene.
We emphasize at this point that, to the best of our knowledge, our system also shows the concept of \textit{event-based deflectometry} for the first time. Naïve event-based deflectometry can be achieved by displaying a moving pattern (point, line, etc.) on an LCD screen, or by projecting the moving pattern on a planar screen or a wall. Such procedures can be seen as a subset of our method, and we already achieve more by using arbitrary scenes of different shapes as screen.

\begin{figure}
    \centering
    \includegraphics[width=0.65\linewidth]{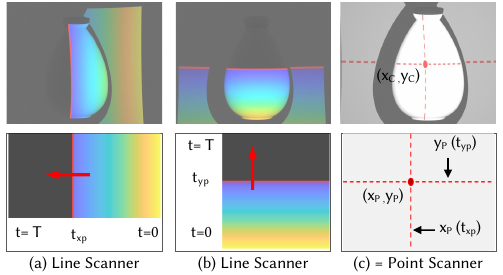}
    \caption{\textbf{Dual Scan Line Projector.} We sweep a line laser horizontally (a) and then vertically (b) to generate events on the camera. To obtain the events for a particular projector ``pixel", we take an intersection of these vertical and horizontal line events (c). This allows us to treat our projection system effectively as a point scanning system, albeit with a much shorter scanning time.}
    \label{fig:DualScanProjector}
\end{figure}

\begin{figure}
    \centering
    \includegraphics[width=0.65\linewidth]{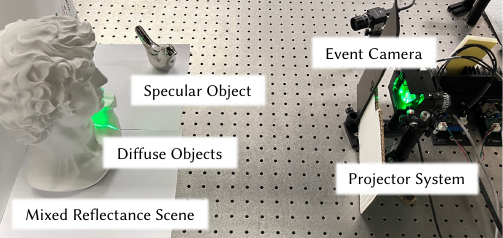}
    \caption{\textbf{Experimental Setup.} Our setup consists of a Prophesee EVK4 event camera and a dual-scanning laser projector. The dual-scanning laser-projector is made from a crosshair (+) laser diode (CivilLaser crosshair laser diode, 80mW, 532nm), a galvo scanning system (Thorlabs GVS012), and a Digital Acquisition Board (NI DAQ PCIe 6363) to control the galvo mirrors.}
    \label{fig:ExperimentalSetup}
\end{figure}

\begin{figure}[b!]
    \input{figures_tex/Fig_teaser_remaining}
    \label{fig:teaser_remaining}
\end{figure}

\section{Results}

\textit{Scene Setup.} We calibrate the scene in a volume of \textit{(height $\times$ width $\times$ depth)} $\approx 25cm \times 45cm \times 60cm$. In this volume, we place different diffuse, specular, and shiny objects and backgrounds (\Cref{fig:ExperimentalSetup}). Our sensor simultaneously scans all surfaces in the scene, and our evaluation procedure calculates their respective shapes.
We test our system against varying ambient lighting conditions (\Cref{fig:VideoFrames}(c)) to ensure the robustness of our method.
Moreover, we quantitatively evaluate the performance of our system for specular and diffuse object reconstructions (\Cref{sec:quantitative}).

\subsection{Imaging Mixed Reflectance Scenes} \label{sec:ImagingMRScenes}

\Cref{fig:more_results,fig:teaser_remaining,fig:teaser} show our results for a variety of scenes.  \Cref{fig:teaser_remaining,fig:more_results} compare our reconstructions against the following state-of-the-art and commercial system baselines: (1) Event-based Structured Light (ESL), (2) Structured light in the presence of global illumination using Ensemble Codes \cite{gupta2011structured}, (3) Intel RealSense D435i\texttrademark \text{ } \cite{intelrealsenseDepthCamera}, and (4) Microsoft Kinect\texttrademark \text{ } v2 Time-of-Flight (ToF) depth sensor \cite{bamji20140, tolgyessy2021evaluation}.
ESL is the current state-of-the-art event-based scanner for diffuse shapes. We compare against the Ensemble Codes \cite{gupta2011structured} method because, although published more than 10 years ago, it still provides state-of-the-art performance of projector-based triangulation systems for shape estimation in the presence of indirect illumination.
The use of ensemble codes suppresses inter-reflections in the scene, providing a high-resolution shape estimation for the diffuse parts of the scene.
We compare against Intel RealSense (triangulation using structured light) and Kinect v2 ToF (time-of-flight) sensors as they are two of the most popular commercial solutions for depth estimation and rely on two fundamentally different measurement principles.
Since our results reach up to 250Hz, the commercial sensors that operate at 30Hz serve as reasonable high-speed comparisons. We chose Kinect v2 as a representative ToF sensor, and although Kinect v3 is available, it still suffers from inherent resolution limitations imposed by CW-ToF (continuous-wave time-of-flight) technology \cite{tolgyessy2021evaluation}.

The Ensemble Codes method \cite{gupta2011structured} requires 50 projector/camera frames to generate one single 3D frame. Although the effective operating speed of the final sensor setup depends on the speed of projector and camera used, it can be said that achieving motion robust performance with this method is very hard.
For the implementation of \cite{gupta2011structured} with our hardware (FLIR BFS-U3-19S4C camera and Viewsonic X11-4K projector), a good SNR in every single camera image was reached for an exposure time of 70ms per frame, resulting in a total capture time ($\tau$) of $50 \times 70ms = 3.5s$ per 3D view.
The Intel RealSense and Kinect v2 ToF were both operated at their default frame rates of 30Hz (33ms per 3D view).

\begin{figure*}
    \centering
    \includegraphics[width=\linewidth]{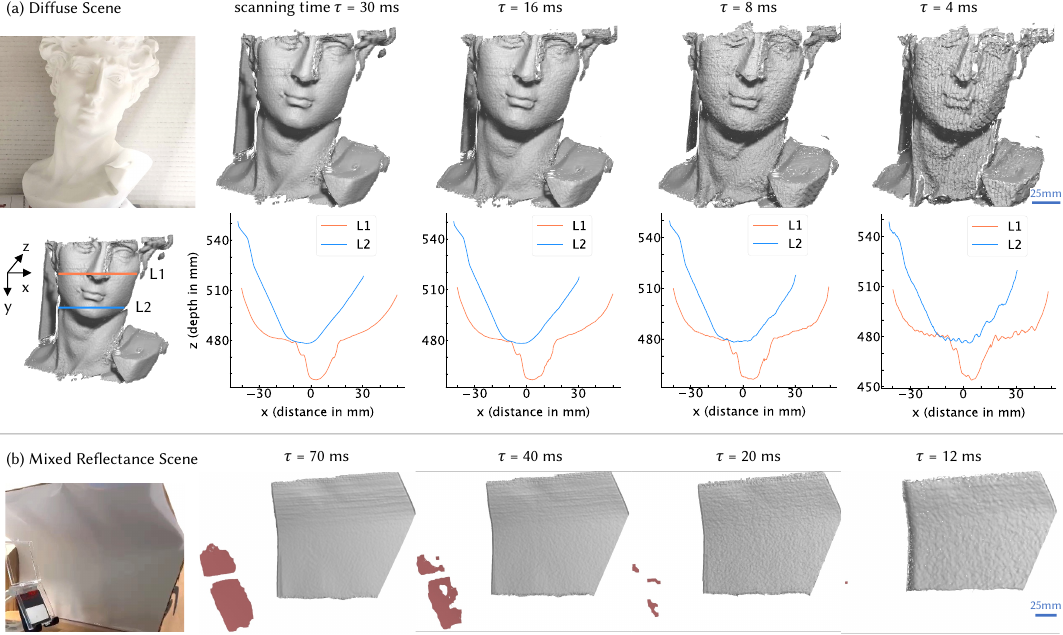}
    \caption{\textbf{Reconstruction quality vs scan time.} (a) For a scene with purely diffuse surfaces (bust) we show our 3D reconstructions for $\tau = 30ms,~16ms,~8ms$, and $4ms$ scans of the scene. Profile plots at two different positions of the bust are shown in the second row. (b)
    Mixed-reflectance scene (planar diffuse surface with partially-specular lens case made of clear plastic), measured at $\tau = 70ms,~40ms, ~20ms$, and $12ms$.
    In both scenes the reconstruction quality of the diffuse parts decreases as we decrease the scan time and the 3d reconstructions display more noise. For (b), faster scan times also result in a reduction in the intensity of secondary light reflections from specular objects, to the point where these changes are not detected by the event camera anymore. This results in a decrease in their coverage and hence the reconstruction quality.}
    \label{fig:QualityVsTime}
\end{figure*}

Our method scans the scene by sweeping a horizontal laser line first, followed by a vertical one. We achieve good SNR values for a 30ms scan in each direction.
We also add a 5ms recovery time between each scan to prevent any probable stray events that might occur during resetting the scanning laser position.
This results in $\tau = (30ms + 5ms) \times 2 = 70ms$ scan time per 3D view, allowing us to do low-noise video capture at a frame rate of about 14Hz.
\Cref{fig:QualityVsTime}(b) empirically shows that a faster scan time reduces the probability of detecting secondary reflections from specular surfaces. In future setup implementations, the scanning time can be further reduced using a laser with higher output power. If the scene only contains diffuse objects (one bounce), our current setup specification allows for a drastic reduction of the scanning time, down to 4ms per 3D view.

Scenes (a) and (b) in \Cref{fig:teaser_remaining}, and Scenes (a)-(c) in \Cref{fig:more_results}, demonstrate our system's ability to image mixed reflectance scenes.
We capture them with a total scanning time of 70ms.
Scene (a) in \Cref{fig:teaser_remaining} consists of a specular object (a convex mirror) and a diffuse object (a bust of Michelangelo's David). As seen in the photo of the scene (first column, laser is on) this scene causes multiple inter-reflections when light is projected on it. Our method is able to reconstruct both the diffuse (in gray) and specular (in blue) parts accurately. Moreover, the reconstructed mirror's radius of curvature is retrieved with high accuracy (\Cref{sec:quantitative}).
The Ensemble Codes method (third column in \Cref{fig:teaser_remaining}, 3.5s scanning time for one 3D view) is not able to reconstruct most of the specular surface and also has some artifacts near David's left eye. Both Ensemble codes and our method reconstruct the diffuse part of the scene with very high quality, with Ensemble Codes slightly beating ours due to the use of a frame-based camera which has four times the number of pixels and has a longer capture time leading to a better signal-to-noise-ratio in comparison.
RealSense and Kinect v2 (fourth and fifth column in \Cref{fig:teaser_remaining}) display some data points at the position of the mirror, but its shape is not detected correctly.
Both commercial sensor concepts reconstruct the diffuse scene part with visibly lower quality.

\begin{figure}
    \input{figures_tex/Fig_more_results}
    \label{fig:more_results}
\end{figure}

\begin{figure}
    \input{figures_tex/Fig_increased_coverage}
    \label{fig:increased_coverage}
\end{figure}

Scene (b) in \Cref{fig:teaser_remaining} consists of a stationary specular mirror, a moving partially specular (shiny) balloon, and several objects with diffuse surfaces: David bust, Stanford bunny, and a moving paper towel. 
It can be seen that our  method performs a high-quality 14Hz reconstruction of the moving scene with surfaces of mixed reflectance
(see videos \href{https://drive.google.com/drive/folders/1l-5CNM5QqyrP5nZwgZ8e4urpwg04lov8?usp=sharing}{\underline{here}})
After scanning, we evaluate the shiny balloon and the convex mirror via deflectometry and the rest of the diffuse objects via triangulation. The result is a \textit{motion-robust} reconstruction of both the balloon and the small mirror surface in the scene along with the diffuse surfaces of Stanford bunny, David statue, and a part of the moving paper towel. 
Ensemble Codes is able to correctly reconstruct the non-moving diffuse scene parts. It misses the moving paper towel, and, as expected, it misses all the specular parts of the scene.
RealSense and Kinect on the other hand are motion robust and perform a lower quality shape estimation of diffuse surfaces as well as the balloon (which is also partially diffuse), but fail in getting the shape of the mirror. 

Similar trends can be observed in \Cref{fig:more_results}:
\Cref{fig:more_results}(a) shows another scene with multiple interreflections caused by a specular ball bearing. Our method uses both the direct reflections from the ball bearing (one bounce specular) as well as the secondary deflectometry reflections to reconstruct the illuminated parts of the ball.
It can be seen that the coverage of the specular ball bearing surface significantly profits from the two screen geometry, which allows capturing the majority of the surface visible to the camera - something that would not be possible with standard deflectometry.
RealSense detects the specular object but fails to accurately reconstruct it, whereas Ensemble Codes and Kinect miss the specular parts.
When scanning the scenes in \Cref{fig:more_results}(b)-(d) with our method, we only illuminated the planar diffuse ``screen" and reconstructed the specular surface solely from 2nd bounce reflections via deflectometry. To our knowledge, this is the first-ever demonstration of \text{Event-Deflectometry}, where a dynamic screen illumination is used in conjunction with an event camera to perform deflectometry measurements of specular object surfaces.

It can be seen that the shapes of the balloon dog (\cref{fig:more_results}(b)) and the metal bird (\cref{fig:more_results}(c)) are reconstructed accurately along with the diffuse reconstructions of the planar screen.
As seen previously, Ensemble Codes misses the specular parts, and RealSense performs a low-quality reconstruction of specular surfaces while Kinect misses them.
For the license plate scene (\cref{fig:more_results}(d)), our method recovers both the diffuse plane and the partially specular license plate. Ensemble Codes misses many regions on the license plate and the diffuse plane, while Realsense and Kinect perform a low-resolution reconstruction of the license plate.

In \Cref{fig:increased_coverage}, we show additional results on mixed reflectance scenes where we reconstruct a full convex mirror in (a), the specular letters F, J and the diffuse letter A (with its fine texture), and a specular telephone object with both the visible sides reconstructed using two different screens in (c).

\subsection{Imaging scenes with purely diffuse surfaces} \label{sec:ImagingDiffuseScenes}

Most event-based 3D scanning literature focuses on scenes with purely diffuse surfaces. Our method also allows for measuring these scenes while providing added degrees of freedom with respect to choosing scanning speed.

We can further reduce our scanning time when working with purely diffuse surfaces. We can perform a single laser line sweep instead of the dual-scanning that is required for deflectometry.
This allows us to reduce the scanning time by 2x without any loss in the reconstruction quality of diffuse scenes. A comparison of our 30ms reconstruction is shown in Scene (c) of \Cref{fig:teaser_remaining}. Our method achieves better data quality than the 3.5s ensemble code scan, and shows significant improvement over that of Kinect and RealSense.
As no (weak) 2-bounce reflections have to be detected anymore for purely diffuse scenes, we can further decrease the time for one single sweep.
In our prototype setup, we reduced the capture time to just 4ms while achieving high-quality reconstructions. A comparison of our ``ultrafast" 4ms-capture with SOTA is shown in the last row (d) of \Cref{fig:teaser_remaining}.  Although slightly lower quality than the 3.5s ensemble code scan and our previous 70ms or 30ms scan, our fast 4ms scan still shows a much higher data quality than RealSense or Kinect (both 33ms). This shows that our method has significant potential for future low-latency 3D scanning implementations - even if only standard diffuse scenes are measured.

In \Cref{fig:QualityVsTime}(a), we show our reconstructions of diffuse surfaces at different scanning speeds, from 30ms to up to 4ms per frame or 250Hz. This is one of the fastest scans shown to date for an event-based structured light system.

\subsection*{Imaging translucent object with high subsurface scattering}

\begin{figure}
    \input{figures_tex/Fig_translucent_objects}
    \label{fig:translucent_objects}
\end{figure}

To further stress‐test our imaging pipeline under extreme scattering conditions,
we reconstruct a  translucent wax statue (see \Cref{fig:translucent_objects}(a)). Pronounced subsurface scattering at the wax surface produces broadened and blurred line‐scan profiles, effectively imposing a low‐pass filter on the recovered geometry. The resulting point cloud is shown in \Cref{fig:translucent_objects}(b). As scattering strength decreases, this filtering effect diminishes, as demonstrated by the high‐fidelity reconstruction of human fingers in \Cref{fig:VideoFrames}(e), where minimal subsurface scattering yields sharply defined features.


\subsection{Comparison with state-of-the-art event-based structured light scanning}
\label{sec:ESL_comparisons}

As discussed, event-based structured light 3D scanning has been previously demonstrated on purely diffuse surfaces \cite{matsuda2015mc3d, muglikar2021esl, wang2020joint, huang2021high, morgenstern2023x}. Herein, we compare our technique to Event-based Structured Light (ESL) \cite{muglikar2021esl}, as this work has demonstrated high speed, high-quality results for diffuse surface reconstruction. For a fair comparison under identical setup geometry, we have reproduced the ESL setup using our event camera (Prophesee EVK4) and a pico projector (Anybeam MEMS laser scanning projector). Triangulation angle, optics, and FoV have been deliberately chosen to be identical to our experiment.
We adopted the ESL pipeline from the authors’ official GitHub repository accompanying \cite{muglikar2021esl}.

ESL reconstructions are shown in \Cref{fig:teaser_remaining} (Column 3), \Cref{fig:more_results} (Column 3), and the raw point clouds are shown in \Cref{fig:esl_point_clouds}. It can be seen that our diffuse surface reconstructions show higher fidelity compared to ESL as seen in \Cref{fig:teaser_remaining}(c, d) and \Cref{fig:esl_point_clouds}(a, b, d). In \Cref{sec:quantitative} we additionally quantify the higher accuracy of our method. ESL is not designed to handle interreflections and specular surfaces, hence fails in such scenes in \Cref{fig:teaser_remaining}(a,b) and \Cref{fig:more_results}.

ESL and our method use the same triangulation principle for diffuse reconstruction.
Two key factors of triangulation performance are system calibration and an accurate estimate of the camera-projector correspondence. 
The camera-projector calibration procedure, often termed as inverse camera calibration, requires establishing x-y correspondences between the camera and the projector by imaging a planar checkerboard calibration target at multiple poses \cite{moreno2012simple}. The fidelity of these correspondences directly governs the calibration accuracy.
Line sweeps generate clean correspondences and can be used for such correspondence.
However, for ESL, the pico projector's fast raster scan approach corresponds to an equivalent line sweep only in the x-direction. Hence, the method relies on gray code calibration, a frame-based camera-projector technique that requires integrating events in some time interval. This leads to x and y correspondences that are highly susceptible to pixel-level event noise, leading to errors in camera projector calibration. Our dual-scanning laser sweep scan (which is operated at slow speeds for acquiring calibration scans) obtains cleaner correspondences and hence reduces calibration errors. For the shown experiments, our obtained camera-projector stereo reprojection error was 0.4 pixels while the error listed in \cite{muglikar2021esl} is 3.6 pixels.

The pico projector implemented in ESL scans in discrete horizontal increments corresponding to its raster pattern. In contrast, our methodology employs continuous scanning, enabling precise sub-pixel correspondences for shape reconstruction. The variational hole-filling strategy adopted by ESL serves to densify the resulting point clouds and could be integrated with our approach to address sparse event distributions. We intentionally omitted this step, as the combination of bias parameters optimization and line-laser sweeping yielded inherently dense event data. Nevertheless, such variational densification would be advantageous for the significantly faster scans (on the order of a few milliseconds). It should be noted, however, that because this variational procedure relies on interpolation rather than the acquisition of new measurements, attenuation of high-frequency spatial information is to be expected.

\begin{figure}
    \input{figures_tex/Fig_esl_point_clouds}
    \label{fig:esl_point_clouds}
\end{figure}


\subsection{Comparison with conventional deflectometry}
\label{sec:Conventional_Deflectometry}

\begin{figure}
    \input{figures_tex/Fig_deflectometry_comparisons.tex}
    \label{fig:deflectometry_comparisons}
\end{figure}

To benchmark our specular shape reconstruction against state-of-the-art deflectometry, we measured each test object using a conventional phase-measuring deflectometry (PMD) setup (see \Cref{fig:deflectometry_comparisons}(a)). We then applied our method utilizing an uncalibrated, movable virtual screen (see \Cref{fig:deflectometry_comparisons}(a-c)) to the same objects. Reconstructions for a convex mirror, a balloon dog, and a bird are presented in \Cref{fig:deflectometry_comparisons}(d-f), respectively. Qualitatively, our uncalibrated virtual screen approach yields results comparable to those obtained with the calibrated PMD system. Furthermore, for both the balloon dog (e) and the bird (f), our method provides increased surface coverage by allowing dynamic adjustment of the screen’s shape and position without necessitating system recalibration. In contrast, conventional deflectometry requires full recalibration of the screen–camera geometry whenever their relative pose is modified.

\subsubsection*{Tradefoff between diffuse surface reflectivity and specular coverage}

Lastly, we also demonstrate the performance of our system for a mixed-reflectance scene as the scan time reduces in \Cref{fig:QualityVsTime}(b). We use a plastic lens case that is less reflective and more transmissive than a mirror.
As the scan becomes faster, the signal intensity for the diffuse bounce and hence its secondary reflection from specular parts becomes lower. The intensity of the second bounce specular reflection eventually becomes lower than the change threshold that the event camera can detect. Hence, its coverage starts vanishing for lower scanning times. Using a laser with higher output power could potentially improve the result.

Similar to \Cref{eq:EventDef}, an event corresponding to a second bounce reflection from a specular surface $S$, denoted by $e^S_{ijt}$ is given by,
\begin{equation}
    e^S_{ijt} = \left(i, j, t, p\right),\text{ }s.t. \text{ }\Delta \log \text{ } I^S_{ijt} > \epsilon,
    \label{eq:SecondEventDef}
\end{equation}
where $\Delta \log \text{ } I^S_{ijt} = |\log(I^S(i,j,t)) - \log(I^S(i,j,t-\Delta t))|$ is governed by
\begin{equation}
    \Delta \log \text{ } I^S_{ijt} \propto \tau \cdot P_L \cdot r_D \cdot r_S.
    \label{eq:EqRefl}
\end{equation}

Here $\tau$ is the scanning time, $P_L$ is the laser power, and $r_D$ and $r_S$ denote the fraction of light reflected from the diffuse and specular surface, respectively. $r_D$ and $r_S$ are a function of the diffuse and specular surface's bidirectional reflectance distribution functions (BRDFs).

Reduced scanning time $\tau$ will lead to a reduced probability of secondary event detection, as can be seen in \Cref{fig:QualityVsTime}(b). A darker reflecting diffuse surface (small $r_D$) would also reflect a lesser fraction of light onto the secondary specular surface, leading to the same effect. This is also true for a smaller reflectivity of the secondary specular surface $r_S$. If it's highly transmissive (like the lens box) or highly absorptive, then the probability of event detection falls proportionally.

\begin{figure*}[t!]
    \centering
    \includegraphics[width=\linewidth]{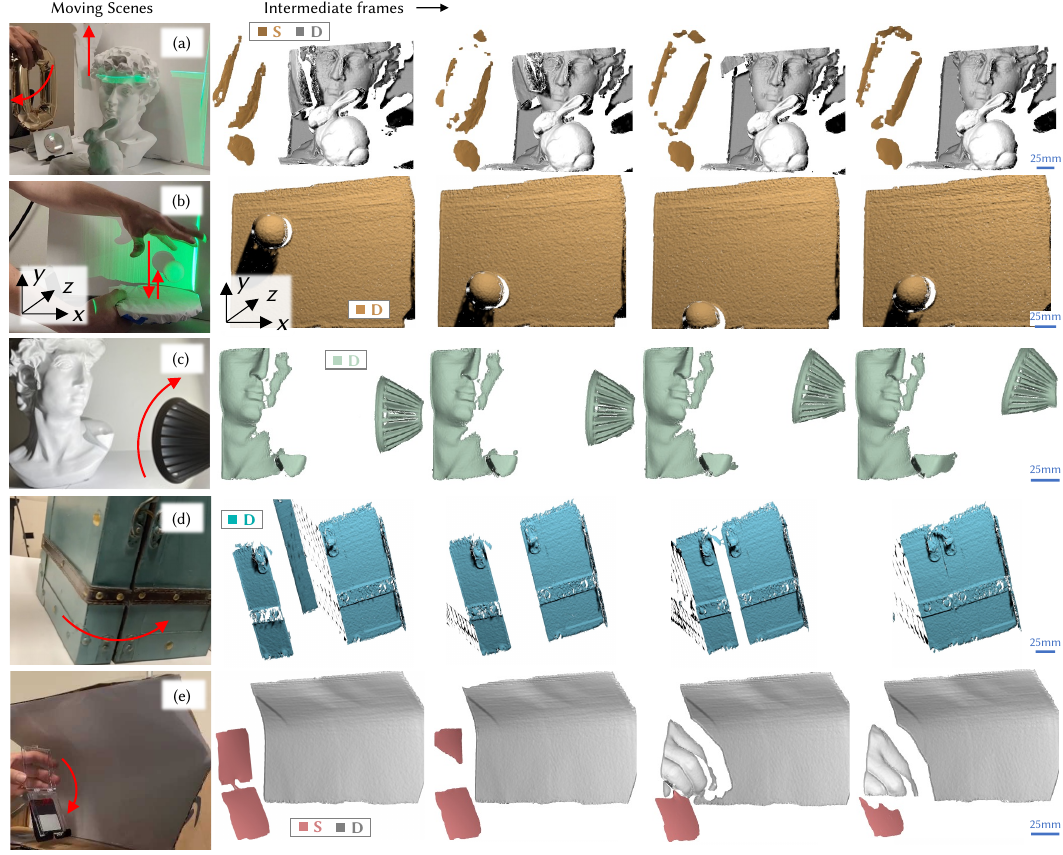}
    \caption{\textbf{Motion-robust 3D measurements of moving scenes} (see videos \href{https://drive.google.com/drive/folders/1l-5CNM5QqyrP5nZwgZ8e4urpwg04lov8?usp=sharing}{\underline{here}}). (a) 14 fps video reconstruction of a mixed reflectance scene containing a specular mirror, a shiny balloon, and other diffuse objects (David bust, Stanford bunny, paper towel). We show intermediate frames from the video where the shiny balloon is rotated and the paper towel is rapidly lifted. (b) 29 fps video reconstruction of a bouncing table tennis ball. (c) 29 fps video reconstruction of a scene with a moving light bulb that causes rapidly changing strong background illumination. Despite the varying ambient light in the scene, we are able to reconstruct the shape of both the bright David bust and the dark side of the light bulb.
   (d) 29 fps video reconstruction of a closing treasure chest that has diffuse as well as partially specular surface components (the metallic lock, and the strip); we are able to reconstruct both surface types using our method.
   (e) 14 fps video reconstruction of closing a partially specular clear plastic lens box by hand where its bottom half is fixed. In the intermediate frames, we can view the transition between seeing the specular top half to later seeing the diffuse fingers as the box is closing.
   (\textit{Notations used}: \textbf{D}: Direct component using triangulation, \textbf{S}: Specular component using deflectometry, Red arrows indicate motion.)}
    \label{fig:VideoFrames}
\end{figure*}

\subsection{Adaptable coverage}

\subsubsection{Multi-camera setup - paving way for $360^\circ$ scene captures}

\textit{Multi-View Capture Using a Second Event Camera.}  

As illustrated in \Cref{fig:2nd_camera_deflectometry}(a), we augment our system with a second event camera, synchronously triggered and calibrated within the overall rig to observe the specular shape from an orthogonal viewpoint, thereby capturing additional reflections of the virtual diffuse screen on the specular shape. The primary camera remains dedicated to imaging the diffuse screen-object itself, while the secondary camera acquires events directly from the specular surface. We then fuse the two reconstructions to generate the final 3D shape. Results for a curved screen and a specular horse model are presented in \Cref{fig:2nd_camera_deflectometry}(b,c), demonstrating proof of concept for a multi-camera configuration for potential extensions toward  $360^\circ$ capture.

\begin{figure}
    \input{figures_tex/Fig_2nd_camera_deflectometry}
    \label{fig:2nd_camera_deflectometry}
\end{figure}

\subsubsection{Moving screen deflectometry}
\label{sec:moving_screen_deflectometry}

In the first-of-its-kind experiment, we employ a handheld planar surface as an uncalibrated, free-hand-guided screen for event-deflectometry measurements. This novel procedure achieves greater coverage of complex specular geometries by aggregating multiple captures rather than relying on a single fixed screen. For each acquisition, the screen is held at a distinct pose, directing reflections onto different regions of the object. When combined, these measurements effectively emulate a large virtual dome-shaped illumination without the need to construct any curved screen. The representative setup and results are shown in \Cref{fig:moving_screen}.

Crucially, unlike conventional deflectometry, this flexible measurement modality does not require recalibration of the screen-camera system after each repositioning: \textit{our event-deflectometry system is self-calibrated}. Although this method is not inherently motion-robust, it clearly demonstrates the system's adaptability. Such a protocol holds considerable promise for rapid industrial inspection and applications demanding versatile coverage, such as assessing hail damage on automotive bodies.

\begin{figure}
    \input{figures_tex/Fig_moving_screen}
    \label{fig:moving_screen}
\end{figure}

\subsection{Quantitative evaluation} \label{sec:quantitative}

We evaluate the accuracy and precision of our method by measuring different objects with known sizes or shapes. For the evaluation of our triangulation reconstruction, we measured a sphere with a diffuse surface and well-known ($1 inch =25.4 mm$) radius, as well as a precisely manufactured diffuse planar surface.
The objects have been placed at a distance of $\sim$60cm to the sensor and the scan time for all our evaluation measurements was 70 ms. To evaluate how accurately our method reconstructs the object shape, we compare our evaluated point cloud with the respective ground truth shape.

For the measurement of the planar surface, we obtain $0.21mm$ as the root-mean-square error (RMSE) with respect to the planar ground truth shape. The RMSE of the sphere data with respect to the spherical ground truth shape is evaluated to $0.31mm$. 
To evaluate the precision of our triangulation reconstruction we isolate the statistical noise on our data by subtracting a low-frequency best-fit surface from our obtained point cloud and eventually evaluate the standard deviation of the remaining 3D noise. The so-obtained precision of the plane measurement is $0.12mm$ and the precision of the sphere measurement is $0.06mm$.

The accuracy and precision of our deflectometry reconstruction are evaluated in a similar fashion, by measuring a specular sphere with a known radius (also $1 inch$).
The diffuse planar surface described above is used as a screen for our deflectometry measurement  (setup similar to the specular ball in Scene a) of \Cref{fig:more_results}).
After automated rejection of outliers in the deflectometry reconstruction outside the $6\sigma$-interval  ($4.8\%$ of data) the RMSE with respect to the specular ground truth shape is calculated to $0.45mm$ and the respective precision to $0.14mm$.

We emphasize that a \textit{deflectometry measurement} with our method always includes a concurrent shape estimation of the diffuse \textit{screen scene parts} (diffuse planar surface in this case). This estimation is also subject to uncertainty (see triangulation error values above), which propagates into the deflectometry measurement.
Our first calculations show that our lateral ``screen position error"  introduced by the our triangulation measurement uncertainty is roughly 10x the lateral back-projection screen error in a well calibrated conventional high-precision deflectometry setup \cite{knauer2004phase}. This explains that our deflectometry accuracy resides in the $\sim 100 \mu m$ range, compared to the single-digit $\mu m$ accuracies (or sometimes even sub-$\mu m$ accuracies) of high-performance deflectometry setups. We also add, however, that this low  error often comes at the cost of limited flexibility and long capture times that often do not allow motion-robust measurements. \\

To better compare the performance of our system to previous event-based structured light methods for diffuse surfaces \cite{muglikar2021esl,matsuda2015mc3d}, we perform an additional accuracy evaluation under slightly different setup geometries. In previous works \cite{muglikar2021esl,matsuda2015mc3d} the system accuracy has been evaluated by comparing the measured absolute distance $d_{est}$ of a planar surface (with respect to the sensor) to the ground truth $d_{gt}$. However, the fact that  $d_{gt}$ (the absolute distance from the sensor to the planar surface) can be barely measured with sufficient  accuracy (in this case: tens of $\mu m$) makes this method unsuitable for error estimation in high-accuracy systems like ours. A suitable related alternative is to perform relative distance measurements of a planar surface that has been translated by a high-precision translation stage. In our experiment, we place a planar surface mounted on a translation stage (Zaber XLHM-100A with a 100 mm range and 0.12 $\mu$m resolution) approximately 500 mm from the camera. The plane at the 50 mm position of the stage is used as reference, and the relative distance is set to 0 mm. Eventually, we move the plane with the translation stage by $\pm 50$ mm in steps of 10 mm.
We first fit a plane to the reference position measurement to obtain its equation parameters. For each plane position \textit{i}, we then translate this reference plane along the translation axis by a known stage displacement $\Delta d_{\mathrm{gt}}^{i}$ to derive the expected plane equation and compute the expected depth $d_{\mathrm{exp}}^{i}(p)$ at each pixel \textit{p} in the region of interest (ROI). Next, we measure the reconstructed depth map $d_{\mathrm{est}}^{i}(p)$ and calculate the root-mean-square error over all \textit{P} pixels in the ROI:
\begin{equation}
\mathrm{RMSE}^{\,i} =
\sqrt{
  \frac{1}{P} \sum_{p=1}^{P} \bigl( d_{\mathrm{est}}^{\,i}(p) - d_{\mathrm{exp}}^{\,i}(p) \bigr)^{2}
}.
\end{equation}

For a fair comparison, we use a 16 ms diffuse scan speed to match ESL's scan speed. \Cref{fig:Accuracy_stage}(b) shows the measured $\Delta d^{i,\mathrm{ref}}_{est}$ as i varies from -50 mm to +50 mm.
The RMSE values for our method remain below 0.6 mm across the entire translation range. For comparison, we perform the same measurement with the ESL setup from \Cref{sec:ESL_comparisons}. The evaluated accuracy values are $\lesssim$ 6 mm.

\textbf{Summary:} For both specular and diffuse reconstructions, the depth error is consistently $< 0.6$ mm, 
over an order of magnitude more accurate than the ESL baseline.
\Cref{tab:Comparison_\thetableA} further shows that our method surpasses prior state-of-the-art techniques, while uniquely enabling mixed-reflectance 3D imaging via a self-calibrated, flexible screen approach.

\begin{figure}
    \input{figures_tex/Fig_accuracy}
    \label{fig:Accuracy_stage}
\end{figure}

\subsection{Motion Robust Capture}

One of the biggest advantages of our system is \textit{fast} high-quality capture which allows for motion robust reconstructions of moving mixed reflectance scenes (\textit{mixed reflectance 3D videos}). Representative frames of these videos for five different scenes are shown in \Cref{fig:VideoFrames}, and the full videos can be seen in the supplementary video.
The first scene (a) shows the reconstruction for the already discussed combination of specular and diffuse objects - a shiny balloon and a convex mirror, and some diffuse objects (David and Stanford Bunny). The moving scene is reconstructed in a 14Hz (70ms per frame) video. The second, third, and fourth scenes (b-d) are reconstructed at 29Hz (35ms per frame), assuming the presence of only diffuse and partially specular object surfaces in the scene. The second scene (b) shows the motion-robust reconstruction of a bouncing table tennis ball, while the third video (c) demonstrates the robustness of our system to (rapidly changing) ambient light:  We move a bright light bulb rapidly, which changes the lighting of the scene.
This still does not affect our evaluation results, leading to the successful reconstruction of the finer features on the side of the bulb (dark region) as well as the David bust (bright region), also demonstrating the high dynamic range of our event-based system.
The fourth scene (d) shows a treasure chest with partially specular surfaces (like the metal strip and handle). Our method can also successfully reconstruct these partially specular parts.
The last scene (e) shows the manual closing of the lens box captured at 14Hz (70ms per frame). In the top section of the lens box, we can see the transition between the top specular flap and the diffuse hand as we close the box. As the hand enters the projector field of view, it reduces the coverage of the specular part and also occludes it from the camera.

\section{Discussion} \label{sec:Discussion}

We introduced a novel pipeline for 3D imaging of mixed reflectance scenes that contain a mix of diffuse, specular, and partially specular object surfaces. Current SOTA methods are not able to measure the entirety of these scenes in a robust, fast, and accurate fashion.
Compared to other approaches, our method is more generalized and shows better performance in terms of dynamic range and motion robustness.
Our dual scanning approach also improves the diffuse reconstruction compared to previous event-based triangulation techniques
and the results are comparable to accurate frame-based camera triangulation setups - an achievement that has not been demonstrated before for event cameras.
We also show the reconstruction of specular surfaces that are indirectly illuminated by the diffuse object screens, eliminating the need for separate deflectometry screens. We believe this approach has the potential to become a widely used high-accuracy, generalized, and motion-robust 3D scanning system in the future.
However, our current approach comes with its own limitations.
As specular surface reconstruction requires a screen for deflectometry, we cannot measure specular surfaces if there are no diffuse surfaces in the scene.

Our algorithm depends on the successful epipolar separation of the direct and indirect components and their further classification into different reflection types (\Cref{fig:AlgorithmExplanation}). In rare cases  (Case VI, type (b) of \Cref{fig:AlgorithmExplanation}), and potentially for complicated specular objects to be measured,  inter-reflections might occasionally fall on the epipolar line of our camera, preventing our algorithm from correctly classifying direct vs. indirect components. Although not observed in our current measurements, this case would lead to a reconstruction artifact. We believe that this rare problem can be solved in future setups by introducing additional hardware-based constraints like polarized illumination
that can resolve the remaining ambiguities.
Unlike traditional deflectometry, where the screen is calibrated with respect to the camera beforehand with extensive calibration, our screen positions are calculated on-the-fly using triangulation. Although this provides great flexibility for our measurements (e.g., \Cref{sec:moving_screen_deflectometry}), it makes the screen positions susceptible to the triangulation error, which then propagates into the deflectometry reconstruction, leading to a reduced accuracy of our deflectometry measurements compared to state-of-the-art setups with frame-based cameras and calibrated screens.

Finally, the coverage of our deflectometry measurements is still limited by the relative positioning and orientation between diffuse screens, specular shapes and the camera position. We emphasize again that these geometric limitations are exactly the same as in screen-based deflectometry, meaning that a setup with identical geometry and identical screen size would achieve identical coverage for both techniques. The big benefit of our method lies in the fact that large and convex screens (e.g., "domes") can be augmented fairly easy and do not require difficult screen calibration. This can lead to interesting potential hardware innovations in the future. For example, a possible direction for future work would be extending the projector system for $360^\circ$ scans in a closed room setting and using a $\sim 360^\circ$ fisheye objective on the event camera. This would turn a closed room into a large dome-like screen, showing an exciting path to exploit the full potential of our method beyond the proof of principle shown in this paper. In the future, we hope that our novel method will contribute to a new wave of computational 3D cameras potentially leading to the effective adoption of our technology, e.g., in AR/VR, medical imaging, or industrial inspection.

\subsection*{Acknowledgements}

This work was supported by NSF award 2153516. We thank Mohit Gupta for providing the code for the Ensemble Codes \cite{gupta2011structured} method which we used for the state-of-the-art comparisons.

\bibliography{references}

\begin{thebibliography}{10}
\newcommand{\enquote}[1]{``#1''}

\bibitem{muglikar2021esl}
M.~Muglikar, G.~Gallego, and D.~Scaramuzza, \enquote{Esl: Event-based structured light,} in \emph{2021 International Conference on 3D Vision (3DV),}  (IEEE, 2021), pp. 1165--1174.

\bibitem{gupta2011structured}
M.~Gupta, A.~Agrawal, A.~Veeraraghavan, and S.~G. Narasimhan, \enquote{Structured light 3d scanning in the presence of global illumination,} in \emph{CVPR 2011,}  (IEEE, 2011), pp. 713--720.

\bibitem{intelrealsenseDepthCamera}
R.~Intel, \enquote{{D}epth {C}amera {D}435i --- intelrealsense.com,} \url{https://www.intelrealsense.com/depth-camera-d435i/} (2023). [Accessed 08-11-2023].

\bibitem{bamji20140}
C.~S. Bamji, P.~O'Connor, T.~Elkhatib, S.~Mehta, B.~Thompson, L.~A. Prather, D.~Snow, O.~C. Akkaya, A.~Daniel, A.~D. Payne \emph{et~al.}, \enquote{A 0.13 $\mu$m cmos system-on-chip for a 512$\times$ 424 time-of-flight image sensor with multi-frequency photo-demodulation up to 130 mhz and 2 gs/s adc,} {\protect\JournalTitle{IEEE Journal of Solid-State Circuits}} \textbf{50}, 303--319 (2014).

\bibitem{o2015homogeneous}
M.~O'Toole, S.~Achar, S.~G. Narasimhan, and K.~N. Kutulakos, \enquote{Homogeneous codes for energy-efficient illumination and imaging,} {\protect\JournalTitle{ACM Transactions on Graphics (ToG)}} \textbf{34}, 1--13 (2015).

\bibitem{gupta2015phasor}
M.~Gupta, S.~K. Nayar, M.~B. Hullin, and J.~Martin, \enquote{Phasor imaging: A generalization of correlation-based time-of-flight imaging,} {\protect\JournalTitle{ACM Transactions on Graphics (ToG)}} \textbf{34}, 1--18 (2015).

\bibitem{nayar2012diffuse}
S.~K. Nayar and M.~Gupta, \enquote{Diffuse structured light,} in \emph{2012 IEEE international conference on computational photography (ICCP),}  (IEEE, 2012), pp. 1--11.

\bibitem{o20143d}
M.~OToole, J.~Mather, and K.~N. Kutulakos, \enquote{3d shape and indirect appearance by structured light transport,} {\protect\JournalTitle{IEEE Transactions on Pattern Analysis and Machine Intelligence}} \textbf{38}, 1298--1312 (2016).

\bibitem{criminisi2005extracting}
A.~Criminisi, S.~B. Kang, R.~Swaminathan, R.~Szeliski, and P.~Anandan, \enquote{Extracting layers and analyzing their specular properties using epipolar-plane-image analysis,} {\protect\JournalTitle{Computer vision and image understanding}} \textbf{97}, 51--85 (2005).

\bibitem{yang2023sepi}
X.~Yang, Q.~Liao, X.~Hu, C.~Shi, and G.~Wang, \enquote{Sepi-3d: soft epipolar 3d shape measurement with an event camera for multipath elimination,} {\protect\JournalTitle{Optics Express}} \textbf{31}, 13328--13341 (2023).

\bibitem{gallego2020event}
G.~Gallego, T.~Delbr{\"u}ck, G.~Orchard, C.~Bartolozzi, B.~Taba, A.~Censi, S.~Leutenegger, A.~J. Davison, J.~Conradt, K.~Daniilidis \emph{et~al.}, \enquote{Event-based vision: A survey,} {\protect\JournalTitle{IEEE transactions on pattern analysis and machine intelligence}} \textbf{44}, 154--180 (2020).

\bibitem{lichtsteiner2008128}
P.~Lichtsteiner, C.~Posch, and T.~Delbruck, \enquote{A 128 $\times$ 128 120 db 15 $\mu$ s latency asynchronous temporal contrast vision sensor,} {\protect\JournalTitle{IEEE journal of solid-state circuits}} \textbf{43}, 566--576 (2008).

\bibitem{brandli2014240}
C.~Brandli, R.~Berner, M.~Yang, S.-C. Liu, and T.~Delbruck, \enquote{A 240$\times$ 180 130 db 3 $\mu$s latency global shutter spatiotemporal vision sensor,} {\protect\JournalTitle{IEEE Journal of Solid-State Circuits}} \textbf{49}, 2333--2341 (2014).

\bibitem{posch2010qvga}
C.~Posch, D.~Matolin, and R.~Wohlgenannt, \enquote{A qvga 143 db dynamic range frame-free pwm image sensor with lossless pixel-level video compression and time-domain cds,} {\protect\JournalTitle{IEEE Journal of Solid-State Circuits}} \textbf{46}, 259--275 (2010).

\bibitem{matsuda2015mc3d}
N.~Matsuda, O.~Cossairt, and M.~Gupta, \enquote{Mc3d: Motion contrast 3d scanning,} in \emph{2015 IEEE International Conference on Computational Photography (ICCP),}  (Houston, TX, USA, 2015), pp. 1--10.

\bibitem{wang2020joint}
Z.~W. Wang, P.~Duan, O.~Cossairt, A.~Katsaggelos, T.~Huang, and B.~Shi, \enquote{Joint filtering of intensity images and neuromorphic events for high-resolution noise-robust imaging,} in \emph{Proceedings of the IEEE/CVF Conference on Computer Vision and Pattern Recognition,}  (2020), pp. 1609--1619.

\bibitem{huang2021high}
X.~Huang, Y.~Zhang, and Z.~Xiong, \enquote{High-speed structured light based 3d scanning using an event camera,} {\protect\JournalTitle{Optics Express}} \textbf{29}, 35864--35876 (2021).

\bibitem{knauer2004phase}
M.~C. Knauer, J.~Kaminski, and G.~Hausler, \enquote{Phase measuring deflectometry: a new approach to measure specular free-form surfaces,} in \emph{Optical Metrology in Production Engineering,}  vol. 5457 (SPIE, 2004), pp. 366--376.

\bibitem{huang2018review}
L.~Huang, M.~Idir, C.~Zuo, and A.~Asundi, \enquote{Review of phase measuring deflectometry,} {\protect\JournalTitle{Optics and Lasers in Engineering}} \textbf{107}, 247--257 (2018).

\bibitem{faber2012deflectometry}
C.~Faber, E.~Olesch, R.~Krobot, and G.~H{\"a}usler, \enquote{{Deflectometry challenges interferometry: the competition gets tougher!}} in \emph{Interferometry XVI: Techniques and Analysis,}  vol. 8493 J.~Schmit, K.~Creath, C.~E. Towers, and J.~Burke, eds., International Society for Optics and Photonics (SPIE, San Diego, CA, USA, 2012), p. 84930R.

\bibitem{hausler2022reflections}
G.~H{\"a}usler and F.~Willomitzer, \enquote{Reflections about the holographic and non-holographic acquisition of surface topography: where are the limits?} {\protect\JournalTitle{Light: Advanced Manufacturing}} \textbf{3}, 226--235 (2022).

\bibitem{woodham1979photometric}
R.~J. Woodham, \enquote{Photometric stereo: A reflectance map technique for determining surface orientation from image intensity,} in \emph{Image understanding systems and industrial applications I,}  vol. 155 (SPIE, 1979), pp. 136--143.

\bibitem{zhang1999shape}
R.~Zhang, P.-S. Tsai, J.~E. Cryer, and M.~Shah, \enquote{Shape-from-shading: a survey,} {\protect\JournalTitle{IEEE transactions on pattern analysis and machine intelligence}} \textbf{21}, 690--706 (1999).

\bibitem{geng2011structured}
J.~Geng, \enquote{Structured-light 3d surface imaging: a tutorial,} {\protect\JournalTitle{Advances in optics and photonics}} \textbf{3}, 128--160 (2011).

\bibitem{zuo2018phase}
C.~Zuo, S.~Feng, L.~Huang, T.~Tao, W.~Yin, and Q.~Chen, \enquote{Phase shifting algorithms for fringe projection profilometry: A review,} {\protect\JournalTitle{Optics and lasers in engineering}} \textbf{109}, 23--59 (2018).

\bibitem{srinivasan1984automated}
V.~Srinivasan, H.-C. Liu, and M.~Halioua, \enquote{Automated phase-measuring profilometry of 3-d diffuse objects,} {\protect\JournalTitle{Applied optics}} \textbf{23}, 3105--3108 (1984).

\bibitem{takeda1983fourier}
M.~Takeda and K.~Mutoh, \enquote{Fourier transform profilometry for the automatic measurement of 3-d object shapes,} {\protect\JournalTitle{Applied optics}} \textbf{22}, 3977--3982 (1983).

\bibitem{willomitzer2017single}
F.~Willomitzer and G.~H{\"a}usler, \enquote{Single-shot 3d motion picture camera with a dense point cloud,} {\protect\JournalTitle{Optics express}} \textbf{25}, 23451--23464 (2017).

\bibitem{willomitzer2013flying}
F.~Willomitzer, S.~Ettl, O.~Arold, and G.~H{\"a}usler, \enquote{Flying triangulation-a motion-robust optical 3d sensor for the real-time shape acquisition of complex objects,} in \emph{AIP Conference Proceedings,}  vol. 1537 (American Institute of Physics, 2013), pp. 19--26.

\bibitem{sundar2022single}
V.~Sundar, S.~Ma, A.~C. Sankaranarayanan, and M.~Gupta, \enquote{Single-photon structured light,} in \emph{Proceedings of the IEEE/CVF Conference on Computer Vision and Pattern Recognition,}  (2022), pp. 17865--17875.

\bibitem{Schaffer:10}
M.~Schaffer, M.~Grosse, and R.~Kowarschik, \enquote{High-speed pattern projection for three-dimensional shape measurement using laser speckles,} {\protect\JournalTitle{Appl. Opt.}} \textbf{49}, 3622--3629 (2010).

\bibitem{mirdehghan2018optimal}
P.~Mirdehghan, W.~Chen, and K.~N. Kutulakos, \enquote{Optimal structured light a la carte,} in \emph{Proceedings of the IEEE Conference on Computer Vision and Pattern Recognition,}  (2018), pp. 6248--6257.

\bibitem{zuo2022deep}
C.~Zuo, J.~Qian, S.~Feng, W.~Yin, Y.~Li, P.~Fan, J.~Han, K.~Qian, and Q.~Chen, \enquote{Deep learning in optical metrology: a review,} {\protect\JournalTitle{Light: Science \& Applications}} \textbf{11}, 39 (2022).

\bibitem{cyberopitcsDSSeries}
CyberOptics, \enquote{Surveyor ds-series, high precision design and accuracy,} \url{https://www.cyberoptics.com/download/industrial-metrology/laser-scanners/DS-SERIES-8026994-REV\_D.pdf} (2023). [Accessed 14-11-2023].

\bibitem{ikeuchi1981determining}
K.~Ikeuchi, \enquote{Determining surface orientations of specular surfaces by using the photometric stereo method,} {\protect\JournalTitle{IEEE Transactions on Pattern Analysis and Machine Intelligence}} pp. 661--669 (1981).

\bibitem{solomon1996extracting}
F.~Solomon and K.~Ikeuchi, \enquote{Extracting the shape and roughness of specular lobe objects using four light photometric stereo,} {\protect\JournalTitle{IEEE Transactions on Pattern Analysis and Machine Intelligence}} \textbf{18}, 449--454 (1996).

\bibitem{rahmann2001reconstruction}
S.~Rahmann and N.~Canterakis, \enquote{Reconstruction of specular surfaces using polarization imaging,} in \emph{Proceedings of the 2001 IEEE Computer Society Conference on Computer Vision and Pattern Recognition. CVPR 2001,}  vol.~1 (IEEE, 2001), pp. I--I.

\bibitem{muglikar2023event}
M.~Muglikar, L.~Bauersfeld, D.~P. Moeys, and D.~Scaramuzza, \enquote{Event-based shape from polarization,} in \emph{Proceedings of the IEEE/CVF Conference on Computer Vision and Pattern Recognition,}  (2023), pp. 1547--1556.

\bibitem{burke2023deflectometry}
J.~Burke, A.~Pak, S.~H{\"o}fer, M.~Ziebarth, M.~Roschani, and J.~Beyerer, \enquote{Deflectometry for specular surfaces: an overview,} {\protect\JournalTitle{Advanced Optical Technologies}} \textbf{12}, 1237687 (2023).

\bibitem{frankot1988method}
R.~T. Frankot and R.~Chellappa, \enquote{A method for enforcing integrability in shape from shading algorithms,} {\protect\JournalTitle{IEEE Transactions on pattern analysis and machine intelligence}} \textbf{10}, 439--451 (1988).

\bibitem{huang2012improvement}
L.~Huang and A.~Asundi, \enquote{Improvement of least-squares integration method with iterative compensations in fringe reflectometry,} {\protect\JournalTitle{Applied optics}} \textbf{51}, 7459--7465 (2012).

\bibitem{huang2015comparison}
L.~Huang, M.~Idir, C.~Zuo, K.~Kaznatcheev, L.~Zhou, and A.~Asundi, \enquote{Comparison of two-dimensional integration methods for shape reconstruction from gradient data,} {\protect\JournalTitle{Optics and Lasers in Engineering}} \textbf{64}, 1--11 (2015).

\bibitem{shimizu2021insight}
Y.~Shimizu, L.-C. Chen, D.~W. Kim, X.~Chen, X.~Li, and H.~Matsukuma, \enquote{An insight into optical metrology in manufacturing,} {\protect\JournalTitle{Measurement Science and Technology}} \textbf{32}, 042003 (2021).

\bibitem{hofer2016infrared}
S.~H{\"o}fer, J.~Burke, and M.~Heizmann, \enquote{Infrared deflectometry for the inspection of diffusely specular surfaces,} {\protect\JournalTitle{Advanced Optical Technologies}} \textbf{5}, 377--387 (2016).

\bibitem{willomitzer2020hand}
F.~Willomitzer, C.-K. Yeh, V.~Gupta, W.~Spies, F.~Schiffers, A.~Katsaggelos, M.~Walton, and O.~Cossairt, \enquote{Hand-guided qualitative deflectometry with a mobile device,} {\protect\JournalTitle{Optics express}} \textbf{28}, 9027--9038 (2020).

\bibitem{liang2016single}
H.~Liang, E.~Olesch, Z.~Yang, and G.~H{\"a}usler, \enquote{Single-shot phase-measuring deflectometry for cornea measurement,} {\protect\JournalTitle{Advanced Optical Technologies}} \textbf{5}, 433--438 (2016).

\bibitem{wang2023accurate}
J.~Wang, T.~Wang, B.~Xu, O.~C. Willomitzer \emph{et~al.}, \enquote{Accurate eye tracking from dense 3d surface reconstructions using single-shot deflectometry,} {\protect\JournalTitle{arXiv preprint arXiv:2308.07298}}  (2023).

\bibitem{wang2023optimization}
T.~Wang, J.~Wang, O.~Cossairt, and F.~Willomitzer, \enquote{Optimization-based eye tracking using deflectometric information,} {\protect\JournalTitle{arXiv preprint arXiv:2303.04997}}  (2023).

\bibitem{oren1997theory}
M.~Oren and S.~K. Nayar, \enquote{A theory of specular surface geometry,} {\protect\JournalTitle{International Journal of Computer Vision}} \textbf{24}, 105--124 (1997).

\bibitem{nishino2004eyes}
K.~Nishino and S.~K. Nayar, \enquote{Eyes for relighting,} {\protect\JournalTitle{ACM Transactions on Graphics (TOG)}} \textbf{23}, 704--711 (2004).

\bibitem{glossyobjects2022}
K.~Tiwary, A.~Dave, N.~Behari, T.~Klinghoffer, A.~Veeraraghavan, and R.~Raskar, \enquote{Orca: Glossy objects as radiance-field cameras,} in \emph{Proceedings of the IEEE/CVF Conference on Computer Vision and Pattern Recognition,}  (2023).

\bibitem{szeliski2022computer}
R.~Szeliski, \emph{Computer vision: algorithms and applications} (Springer Nature, 2022).

\bibitem{tsai2016shape}
C.-Y. Tsai, A.~Veeraraghavan, and A.~C. Sankaranarayanan, \enquote{Shape and reflectance from two-bounce light transients,} in \emph{2016 IEEE International Conference on Computational Photography (ICCP),}  (IEEE, 2016), pp. 1--10.

\bibitem{velten2012recovering}
A.~Velten, T.~Willwacher, O.~Gupta, A.~Veeraraghavan, M.~G. Bawendi, and R.~Raskar, \enquote{Recovering three-dimensional shape around a corner using ultrafast time-of-flight imaging,} {\protect\JournalTitle{Nature communications}} \textbf{3}, 745 (2012).

\bibitem{heide2014diffuse}
F.~Heide, L.~Xiao, W.~Heidrich, and M.~B. Hullin, \enquote{Diffuse mirrors: 3d reconstruction from diffuse indirect illumination using inexpensive time-of-flight sensors,} in \emph{Proceedings of the IEEE Conference on Computer Vision and Pattern Recognition,}  (2014), pp. 3222--3229.

\bibitem{huang2011study}
L.~Huang and A.~Asundi, \enquote{Study on three-dimensional shape measurement of partially diffuse and specular reflective surfaces with fringe projection technique and fringe reflection technique,} in \emph{Dimensional Optical Metrology and Inspection for Practical Applications,}  vol. 8133 (SPIE, 2011), pp. 24--30.

\bibitem{liu20203d}
X.~Liu, Z.~Zhang, N.~Gao, and Z.~Meng, \enquote{3d shape measurement of diffused/specular surface by combining fringe projection and direct phase measuring deflectometry,} {\protect\JournalTitle{Optics Express}} \textbf{28}, 27561--27574 (2020).

\bibitem{karami2022combining}
A.~Karami, F.~Menna, and F.~Remondino, \enquote{Combining photogrammetry and photometric stereo to achieve precise and complete 3d reconstruction,} {\protect\JournalTitle{Sensors}} \textbf{22}, 8172 (2022).

\bibitem{morgenstern2023x}
W.~Morgenstern, N.~Gard, S.~Baumann, A.~Hilsmann, and P.~Eisert, \enquote{X-maps: Direct depth lookup for event-based structured light systems,} in \emph{Proceedings of the IEEE/CVF Conference on Computer Vision and Pattern Recognition,}  (2023), pp. 4006--4014.

\bibitem{raskar1998office}
R.~Raskar, G.~Welch, M.~Cutts, A.~Lake, L.~Stesin, and H.~Fuchs, \enquote{The office of the future: A unified approach to image-based modeling and spatially immersive displays,} in \emph{Proceedings of the 25th annual conference on Computer graphics and interactive techniques,}  (1998), pp. 179--188.

\bibitem{raskar2002projector}
R.~Raskar, \enquote{Projector-based three dimensional graphics,} Ph.D. thesis, University of North Carolina at Chapel Hill (2002).

\bibitem{xin2019theory}
S.~Xin, S.~Nousias, K.~N. Kutulakos, A.~C. Sankaranarayanan, S.~G. Narasimhan, and I.~Gkioulekas, \enquote{A theory of fermat paths for non-line-of-sight shape reconstruction,} in \emph{Proceedings of the IEEE/CVF conference on computer vision and pattern recognition,}  (2019), pp. 6800--6809.

\bibitem{faccio2020non}
D.~Faccio, A.~Velten, and G.~Wetzstein, \enquote{Non-line-of-sight imaging,} {\protect\JournalTitle{Nature Reviews Physics}} \textbf{2}, 318--327 (2020).

\bibitem{willomitzer2021fast}
F.~Willomitzer, P.~V. Rangarajan, F.~Li, M.~M. Balaji, M.~P. Christensen, and O.~Cossairt, \enquote{Fast non-line-of-sight imaging with high-resolution and wide field of view using synthetic wavelength holography,} {\protect\JournalTitle{Nature communications}} \textbf{12}, 6647 (2021).

\bibitem{gu2023fast}
C.~Gu, T.~Sultan, K.~Masumnia-Bisheh, L.~Waller, and A.~Velten, \enquote{Fast non-line-of-sight imaging with non-planar relay surfaces,} in \emph{2023 IEEE International Conference on Computational Photography (ICCP),}  (IEEE, 2023), pp. 1--12.

\bibitem{tolgyessy2021evaluation}
M.~T{\"o}lgyessy, M.~Dekan, L.~Chovanec, and P.~Hubinsk{\`y}, \enquote{Evaluation of the azure kinect and its comparison to kinect v1 and kinect v2,} {\protect\JournalTitle{Sensors}} \textbf{21}, 413 (2021).

\bibitem{moreno2012simple}
D.~Moreno and G.~Taubin, \enquote{Simple, accurate, and robust projector-camera calibration,} in \emph{2012 Second International Conference on 3D Imaging, Modeling, Processing, Visualization \& Transmission,}  (IEEE, 2012), pp. 464--471.

\end{thebibliography}

\end{document}